\documentclass[journal]{IEEEtran}
\usepackage{graphicx}
\usepackage{subfigure}  
\usepackage[ruled]{algorithm2e}
\usepackage{multirow} 
\usepackage{amsmath} 
\usepackage{cite}
\usepackage{booktabs}
\usepackage{footnote}
\usepackage{makecell}
\usepackage{float}
\ifCLASSINFOpdf
\else
\fi
\begin{document}
%
\title{
  A Multi-intersection Vehicular Cooperative Control based on End-Edge-Cloud Computing
  }
%
%
%
\author{
  \IEEEauthorblockN{
    Mingzhi Jiang,
    Tianhao Wu,
    Zhe Wang,
    Yi Gong,\\
    Lin Zhang*, \IEEEmembership{Member, IEEE}, 
    Ren Ping Liu, \IEEEmembership{Senior Member, IEEE}
  }
  




\thanks{M. Jiang, T. Wu and Z. Wang are with the School of Artificial Intelligence, Beijing University of Posts and Telecommunications, Beijing 100876 China.}
\thanks{Y. Gong is with the School of Information and Communication Engineering, Beijing Information Science and Technology University, Beijing 100192 China.}
\thanks{L. Zhang is with Beijing Information Science and Technology University, Beijing 100192 China. He is also with Beijing University of Posts and Telecommunications, Beijing 100876 China.(e-mail:zhl@bistu.edu.cn)}
\thanks{R. P. Liu is with University of Technology Sydney (UTS), Global Big Data Technologies Center (GBDTC), Sydney, Australia.}
\thanks{Manuscript received XX XX, XXXX; revised XX XX, XXXX.}}

%
%

\markboth{IEEE Transactions on Vehicular Technology,~Vol.~XX, No.~XX, XX~XXXX}%
{Mingzhi \MakeLowercase{\textit{et al.}}: A Reinforcement Learning-based Cooperative Control Scheme for Vehicle End-Edge-Cloud Computing at Large-scale Unsignalized Intersections}
%



\maketitle

\begin{abstract}
  Cooperative Intelligent Transportation Systems (C-ITS) will change the modes of road safety and traffic management, especially at intersections without traffic lights, namely unsignalized intersections.
  Existing researches focus on vehicle control within a small area around an unsignalized intersection. 
  In this paper, we expand the control domain to a large area with multiple intersections.
  In particular, we propose a Multi-intersection Vehicular Cooperative Control (MiVeCC) to enable cooperation among vehicles in a large area with multiple unsignalized intersections.
  Firstly, a vehicular end-edge-cloud computing framework is proposed to facilitate end-edge-cloud vertical cooperation and horizontal cooperation among vehicles.
  Then, the vehicular cooperative control problems in the cloud and edge layers are formulated as Markov Decision Process (MDP) and solved by two-stage reinforcement learning. 
  Furthermore, to deal with high-density traffic, vehicle selection methods are proposed to reduce the state space and accelerate algorithm convergence without performance degradation.
  A multi-intersection simulation platform is developed to evaluate the proposed scheme. Simulation results show that the proposed MiVeCC can improve travel efficiency at multiple intersections by up to 4.59 times without collision compared with existing methods.
\end{abstract}


%
\IEEEpeerreviewmaketitle

\section{Introduction}
%
%
%
%

\IEEEPARstart{T}{he} growing demand for mobility puts more pressure on the transportation system than ever before, which causes terrible traffic congestion and inefficient transportation.
The development of wireless communication technology, especially vehicular communication technology, gives vehicles the ability to exchange real-time information with pedestrians, vehicles, roadside infrastructures, and the cloud. 
Meanwhile, the maturity of Artificial Intelligence (AI) technology improves vehicles' ability to process complex road information. 
Cooperative Intelligent Transportation Systems (C-ITS)\cite{DBLP:conf/itsc/SeredynskiV16}, as one of the representatives of the integration of advanced wireless communication and AI technologies, is considered a promising solution to alleviate traffic congestion. 
Recently, many studies have been conducted on cooperative control of vehicles at intersections based on C-ITS.

\begin{figure*}[ht]
  \centering
  \includegraphics[width=\linewidth]{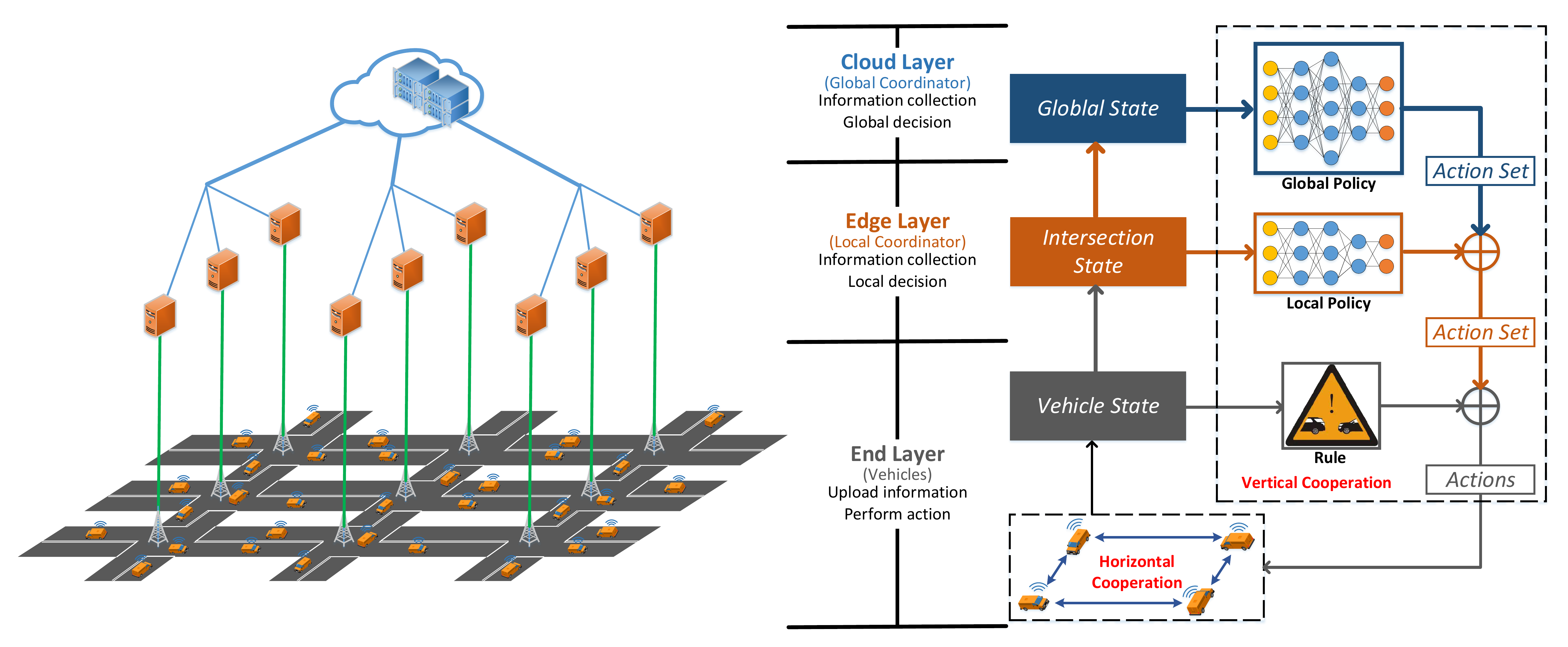}
  \caption{The framework of MiVeCC with end-edge-cloud computing architecture. In this framework, vehicle, local coordinator, and global coordinator correspond to end node, edge node, and cloud node. The framework contains end-edge-cloud vertical cooperation and horizontal cooperation among vehicles.
  The actions performed by vehicles are the combination of the two cooperation. }
  \label{fig:general_architecture}
  \end{figure*}

Vehicle cooperative control at intersections can be classified into two categories, signalized and unsignalized.
Traditionally, under signalized control, vehicles passing through the intersection are guided by traffic lights in an ordered manner. Researchers utilized various methods to optimize the phase and duration of traffic lights, such as convex optimization\cite{DBLP:journals/tcst/GrandinettiWG19}, multi-agent\cite{DBLP:journals/tits/WuLX19}, Petri Nets\cite{DBLP:journals/tits/LuoHW20}, consortium blockchain\cite{DBLP:journals/access/ZhangW19d}, and evolutionary computation\cite{DBLP:journals/access/SegredoLSA19}. 
However, resulting from the start-up and clearance loss of vehicles at intersections, signalized control methods are believed to become the performance bottleneck for connected and automated driving\cite{DBLP:journals/iotj/QianZLLMH19}.

Unsignalized control methods are enabled by autonomous driving and connected vehicle. These methods have the potential to eliminate traffic congestion and significantly improve transport efficiency. 
There are two main types of unsignalized control, i.e. negotiation-based schemes\cite{tachet2016revisiting,levin2016paradoxes} and planning-based schemes\cite{DBLP:conf/amcc/ZhangMC16}.
Planning-based schemes can achieve better performance but require more computation, failing to do so may incur performance penalties. 
As such, new methods, such as Reinforcement Learning (RL) have been introduced to optimized unsignalized intersection control.
RL endows the control scheme with better performance due to its self-evolution ability. For example, in singlaized control, researchers used DQN\cite{DBLP:journals/tvt/LiangDWH19} and LSTM\cite{DBLP:journals/access/LiaoLWZPSLZ20} to optimize traffic light durations. In unsignalized control schemes, researchers utilized the variation of PPO \cite{DBLP:journals/tvt/GuanRLSLL20} to dispatch vehicles passing through intersections. 
Until now, RL-based unsignalized methods are limited to the single intersection.

Coordinating vehicles across multiple unsignalized intersections could bring additional benefits, but also raise major challenges, in terms of computing power, control range and delay.
Generally, the end-edge-cloud architecture includes three layers of tasks\cite{qiu2020edge}. The end layer collects data and performs actions, the edge layer is responsible for processing subtasks, and the cloud layer takes charge of data mining and resource allocation. 
This paper introduces an end-edge-cloud architecture into intersection control to improve driving efficiency by coordinating tasks at different layers. 
To the best of our knowledge, this is the first attempt to apply end-edge-cloud architecture to vehicular cooperative control at multiple unsignalized intersections. 

In this paper, a Multi-intersection Vehicular Cooperative Control (MiVeCC) is proposed based on end-edge-cloud computing. The general architecture is presented in Fig.\ref{fig:general_architecture}.

The main contributions of this paper are four-fold.
\begin{itemize}
  \item By considering multi-layer decision-making, followed by optimizing vehicular cooperative control globally, a novel vehicular end-edge-cloud computing framework is proposed to implement end-edge-cloud vertical cooperation and horizontal cooperation among vehicles.
  \item To optimize vehicular cooperative control across multiple intersections, a two-stage reinforcement learning is adopted after formulating the vehicle control problem as Markov Decision Process (MDP).
  \item Vehicle selection methods are proposed to reduce the state space and accelerate algorithm convergence without performance degradation.
  \item Extensive simulations are conducted on a newly developed platform, which incorporates unsignalized intersections and vehicle control. The results demonstrate that, compared with the benchmark scheme, the proposed scheme can improve the travel efficiency at multiple intersections by up to 4.59 times. 
\end{itemize}

The remaining parts of this paper are organized as follows. 
Section II reviews some related works on unsignalized intersection control.
Section III proposes a vehicle end-edge-cloud computing framework to achieve horizontal and vertical cooperation.
Section VI formulates the vehicle control as MDP problems and adopts a two-stage reinforcement learning.
A vehicle selection method is presented to cope with high-density traffic in Section VI.
Section VII elaborates the details of the algorithm architecture used in the proposed scheme.
Section VII provides simulation results for the performance of the proposed scheme.
Lastly, in Section VIII, the conclusions are presented and some future directions are indicated.

\section{Related Works}
The intersection control schemes can be classified into signalized schemes and unsignalized schemes.
The signalized intersection control schemes have been widely deployed and deeply studied.
Gao et al.\cite{DBLP:journals/tits/GaoZZSS19} utilized the harmony search and artificial bee colony to address a bio-objective urban traffic light scheduling problem.
Except the method of optimizing the phase and duration of traffic lights, the method of changing the vehicle's driving behavior is another solution.
Mofan et al.\cite{DBLP:journals/tits/ZhouYQ20} proposed an RL-based car-following model for connected and automated vehicles to obtain an appropriate driving behavior to improve travel efficiency, fuel consumption and safety at signalized intersections in real-time.
The signalized methods can be extended to applying in the scenario of multiple intersections. However, the increasing number of observations and actions will bring a high computation.
Ge et al.\cite{DBLP:journals/access/GeSWRT19} adopted a distributed control method, where each intersection interacts with adjacent intersections by Q-value transfer.
Tan et al.\cite{DBLP:journals/tcyb/TanBDJDW20} trained the separate base model for every single intersection and add a global dense layer for cooperation among intersections.
The above methods under the control of traffic lights are all defined as the phase-switching mechanism. However, this mechanism will lead to a time loss due to traffic's behavior with start and stop frequently.

The unsignalized methods have become a hot research topic at intersections to improve traffic efficiency.
Generally, there are two types of unsignalized methods, i.e., distributed and centralized.
The Distributed control method can be realized without additional roadside units, which means that communication between vehicles plays a major role.
Xu et al.\cite{xu2018distributed} used a spanning tree to redefine the relationship between vehicles, where the vehicles corresponding to the same layer can pass the intersection without collision at the same time.
Based on the virtual lane, Wu et al.\cite{wu2020cooperative} decoupled the relationship between the identity and driving information of vehicles and proposed a cooperative RL method, named CoMADDPG. The cooperative RL method adjusted the vehicle actions to pass the intersection safely and effectively.
Bian et al. presented a task-area framework to decompose the mission of cooperative passing into three tasks, including state observation, arriving time optimization and trajectory tracking control\cite{DBLP:journals/tie/BianLRWLL20}.
To consider more factors for optimizing vehicle passing at intersections, researchers treat centralized control methods as an indispensable category.
Muhammed et al.\cite{DBLP:journals/tits/SayinLSSB19} considered the traffic priority of vehicles at intersections from the perspective of the number of passengers, and proposed a compensation-based incentive and compatible mechanism to avoid false information reported by vehicles.
Zhang et al.\cite{zhang2016optimal} used real-time planning to ensure that vehicles safe passing at intersections while minimizing fuel consumption.
Unfortunately, almost all the unsignalized intersection control methods are limited to the single-intersection scenario. However, centralized control schemes can provide the basic idea for extending the scenario to the multiple-intersection.

Onboard computing (i.e. computing on the end node) can produce the fastest response, but the computing capacity is not adequate to afford high complexity algorithms. 
Cloud computing can gather more information for accurate decisions, but the communication latency is high.
Edge computing is a trade-off between the above two. 
Any single computing mode mentioned above cannot cope with the diversity of task objectives.
Therefore, the collaboration among end, edge and cloud is a promising direction.
Ren et al.\cite{DBLP:journals/tvt/RenYHL19} derived an optimal task splitting strategy for a joint communication and computation resource problem. In the strategy, the tasks of end nodes can be partially processed at the edge and cloud node.
Liu et al.\cite{DBLP:journals/tvt/LiuZCHG20} presented an example of unmanned-aerial-vehicle swarms computation offloading in end-edge-cloud computing paradigm and proposed an online algorithm for jointly optimization of the computation offloading and multi-hop routing scheduling.
However, existing research achievements focus on the computing offloading of abstract tasks, and there is little research on end-edge-cloud cooperation for specific tasks, such as vehicle cooperative control at intersections.

To verify the control schemes' performance of autonomous driving, simulation platforms are investigated, such as \cite{DBLP:journals/tits/SayinLSSB19, DBLP:conf/vtc/LuK18, DBLP:journals/tac/AhnV18} for the single intersection and \cite{DBLP:journals/tits/WegVHH19, DBLP:conf/itsc/WangWWZZ19, DBLP:conf/iros/HuSG19} for the multiple intersections. 
However, these simulation platforms are not suitable for control schemes in a large area with multiple intersections. 
Therefore, a new simulation platform needs to be designed.

\section{A Vehicle End-edge-cloud Computing framework}

This paper proposes a vehicular end-edge-cloud computing framework for cooperative control at unsignalized intersections. There are three layers (end, edge, and cloud).
The entire framework can refer to Fig.\ref{fig:general_architecture}.

In the end layer, vehicles act as end nodes to interact with other vehicles and upper layers (cloud layer and edge layer). 
Vehicles perform the actions received from the upper layers. These actions are constrained by the onboard collision avoidance rules.
Thus, the actions performed by vehicles are the combination of the output of the three layers.
While performing actions, vehicles collect driving information (position, velocity, acceleration, etc.) to upload them to upper layers and share with adjacent vehicles.

In the edge and cloud layer, nodes have similar functions, including vehicle information collection and decision-making, but differ in control range and objective.
Edge nodes act as local coordinators to control vehicles within the corresponding single intersection.
The cloud node acts as the global coordinator to control vehicles with a global policy in the entire road network.
Regarding the objectives, edge nodes focus on safety and efficiency with local policy, while the cloud node majors in traffic density adjustment to further improve efficiency.

The framework's goal is to construct vertical and horizontal cooperation, which is the source of actions to be performed by vehicles. 
Vertical collaboration means optimizing vehicles' actions with the help of end-edge-cloud structures.
Horizontal collaboration means avoiding the collision by sharing driving information among neighboring vehicles to achieve safe driving.

\section{Problem Statement and Reinforcement Learning Formulation}

\subsection{Problem Statement}

This paper considers an unsignalized traffic control system composed of multiple intersections $\textbf{I}=\{ {I_i},i \in N^{+}\}$. As shown in Fig.\ref{fig:general_architecture}, each intersection consists of four-way single-lane. There is an edge coordinator, denoted by $E_i$, near each intersection, which can communicate with vehicles and cloud coordinator $C$. At each time step, vehicle set near intersection $I_i$ is denoted as $V_{edge_i}$. The cloud coordinator sends instructions, indicated by action set $A_{edge}$, to all vehicles, while the edge coordinator sends instructions, denoted by action set $A_{edge}$, to the vehicles in its control area. The vehicles are instructed by different edge coordinators in turns. The longitudinal motion of vehicles is given by
\begin{equation}
    \begin{aligned}
      x^{long}(t + 1) & = x^{long}(t) + v(t)T  + \frac{1}{2}a(t)T^2 \\
      v(t + 1) & = v(t) + a(t)T
      \end{aligned},
    \label{eq:x_v_t}
\end{equation}
where $x^{long}$ is the displacement, $v$ and $a$ are the velocity and acceleration respectively, and $T$ is the discrete-time step.
It can be found that the change of vehicle motion state depends on the change of acceleration at the previous time step. 
Since all vehicles have the above characteristics, multi-vehicle control at unsignalized intersections can be modeled as an MDP problem.

Same as related work\cite{DBLP:journals/tie/BianLRWLL20}, there are four assumptions to support our work. 
\begin{itemize}
  \item This work focus on the problem of longitudinal vehicle control. Therefore, all vehicles keep their original directions and go straight when they arrived at the intersection.
  \item All vehicles are connected and automated, which can measure kinetic information and communicate with coordinators and adjacent vehicles.
  \item Considering the simplification of the problem, communication latency and package loss are not taken into account in this paper.
  \item The entry time of vehicles follows the Poisson process, which is a commonly used assumption.
\end{itemize}

To maximize the overall traffic efficiency in the system, two kinds of coordinators are designed. They are edge coordinators and the cloud coordinator. Each coordinator is capable of vehicle control within their scopes. Each coordinator corresponds to an MDP.
$\{x,u,r,p\}$ is a typical four-tuple for MDP. $x$,$u$,$r$, and $p$ stand for state, action, reward and transition model, respectively.
In this paper, two MDPs are proposed to solve the vehicle control problem at unsignalized intersections. 
There are two kinds of MDP, $\{x_{E}(t),u_{E}(t),r_{E}(t),p_{E}\}$ and $\{x_{C}(t),u_{C}(t),r_{C}(t),p_{C}\}$ respectively for edge node and cloud node.
The state transition in the above two kinds of nodes can be described as follows,
\begin{equation}
  x(t+1)\sim p(x(t),u(t),z(t)),
\end{equation}
where $z(t)$ stands for the interaction factor, which is used to take the decision-making from other layers. For example, the transition model of the edge node is influenced by both the cloud node and end nodes. Besides, the cloud node is influenced by both edge nodes and end nodes.

Additionally, the vehicles integrate information from different layers, and output the final action.
From the perspective of each layer, the functions and control scope are different.
The cloud provides whole traffic guidance in a large area with multiple intersections. 
The edge is responsible for vehicle control at the corresponding intersection to alleviate local traffic congestion.
The end, which is the vehicle, is only responsible for itself, providing the last defense line to collisions.

\subsection{Reinforcement Learning Formulation}

To address the MDP problems mentioned above, a two-stage RL algorithm is adopted in our work. Stage-I RL runs on the edge nodes, and Stage-II RL runs on the cloud node. The two stages of RL are interacted by the action performed by vehicles.
At first, this paper utilizes two concepts, decision-vehicles and reference-vehicles. 
Decision-vehicles are vehicles whose state can be changed at the current time step. 
Reference-vehicles are vehicles that provide driving information to assist decision-vehicle in changing its state. For a specific physical vehicle, both concepts can exist simultaneously.
As the three essential RL elements (state, action and reward) are described as following.

\begin{figure}
  \centering
  \includegraphics[width=\linewidth]{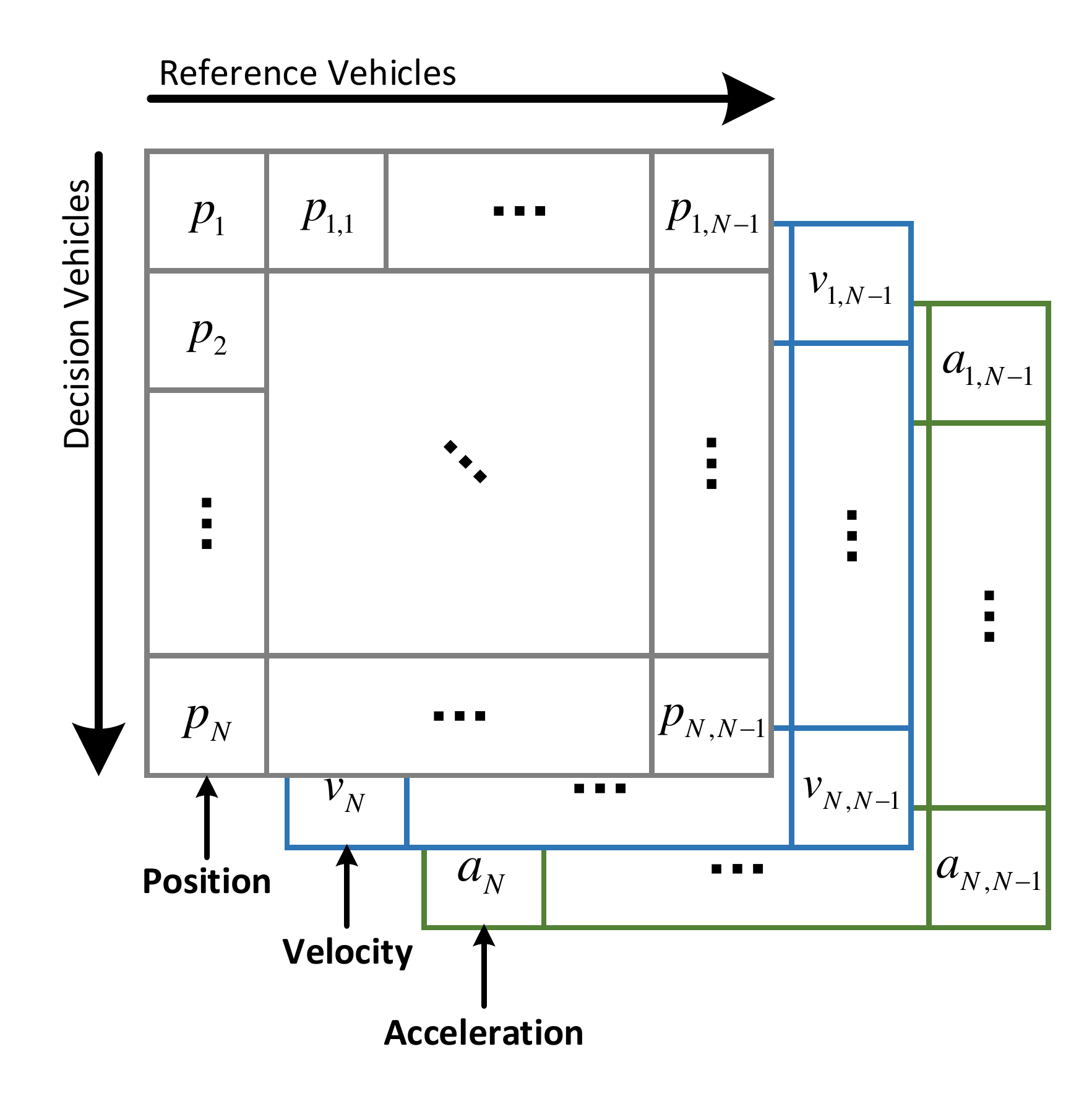}
  \caption{State Graph. The cloud node and edge nodes collect vehicle kinetic information and construct three state graphs for position, velocity and acceleration. Firstly, the nodes select decision-vehicles as the first row of the graph. Then, the nodes select the reference-vehicles as each row of the graph for the decision-vehicles according to the nearest vehicle principle.}
  \label{fig:state_graph} 
\end{figure}

In the real intersection scenario, the number of vehicles is uncertain. 
Therefore, the key to cooperative control at intersections is to convert uncertainty into certainty under some specific scenarios. This paper uses vehicle selection to complete this conversion, which is described in Section \ref{sec:vehicle_selection}.
The edge nodes collect the kinetic information (position, velocity, and acceleration) from vehicles via vehicle-to-infrastructure communication, and select decision-vehicles with the highest collision probability with others. The collision probability is described by safety value (SV), which is defined in Section \ref{sec:safety_value}.
The cloud node collects the kinetic information from edge nodes, and selects decision-vehicles with the density. High density represents high collision probability and low density means a low traffic efficiency, which is explained in Section \ref{sec:density}.
With collected vehicle kinetic information, State Graph (SG) is produced by the cloud node and edge nodes, which is shown in Fig.\ref{fig:state_graph}. 
Firstly, the nodes select decision-vehicles as the first row of the graph. 
Then, the nodes select the reference-vehicles as a row of the graph for the decision-vehicles according to the nearest vehicle principle. 
For the subsequent convolution operation, the maximum of reference-vehicles is one less than the maximum of decision-vehicles to ensure that the SG is square.
The difference between the cloud and the edge in the state is the size of data, and the specific value is explained in the following experiment part.

After the status information is transmitted to the policy module at edge/cloud nodes, the output results will guide the decision-vehicle to move.
Here, acceleration is selected as the action, and the acceleration value is continuous with maximum and minimum.
It is worth noting that the output actions form a set whose position is exactly the column vector of SG.

Since the background of the problem is a continuous flow of traffic, the reward is not defined with the final situation, such as vehicle passing or collision but scattered in the running process. This paper uses velocity as the reward for supporting the training process of RL.

\section{Vehicle Selection Methods Based on Multiple Criteria}
\label{sec:vehicle_selection}

Neural Network (NN) is adopted as the nonlinear mapping from state to action.
To reduce the size of network structure and accelerate convergence without performance degradation, the mechanism of vehicle selection is introduced to limit the input and output size of NN.
Until now, this sub-problem has turned into how to select vehicles including types of decision-vehicle and reference-vehicle to construct the state effectively.

The decision-vehicles are the vehicles that will receive a new action instruction at this time step, while the reference-vehicles are selected to assist the instruction production for decision-vehicles. Therefore, the selection of decision-vehicles should consider both traffic safety and efficiency, and the selection of reference-vehicles should consider their effect on decision-vehicles.
In this paper, vehicle selection methods are proposed based on multiple criteria. 
The selection of decision-vehicles depends on both SV and density separately.
The selection of reference-vehicles depends on the distance on the virtual lane.

\subsection{Distance on the Virtual Lane}

\begin{figure}
  \centering
  \includegraphics[width=\linewidth]{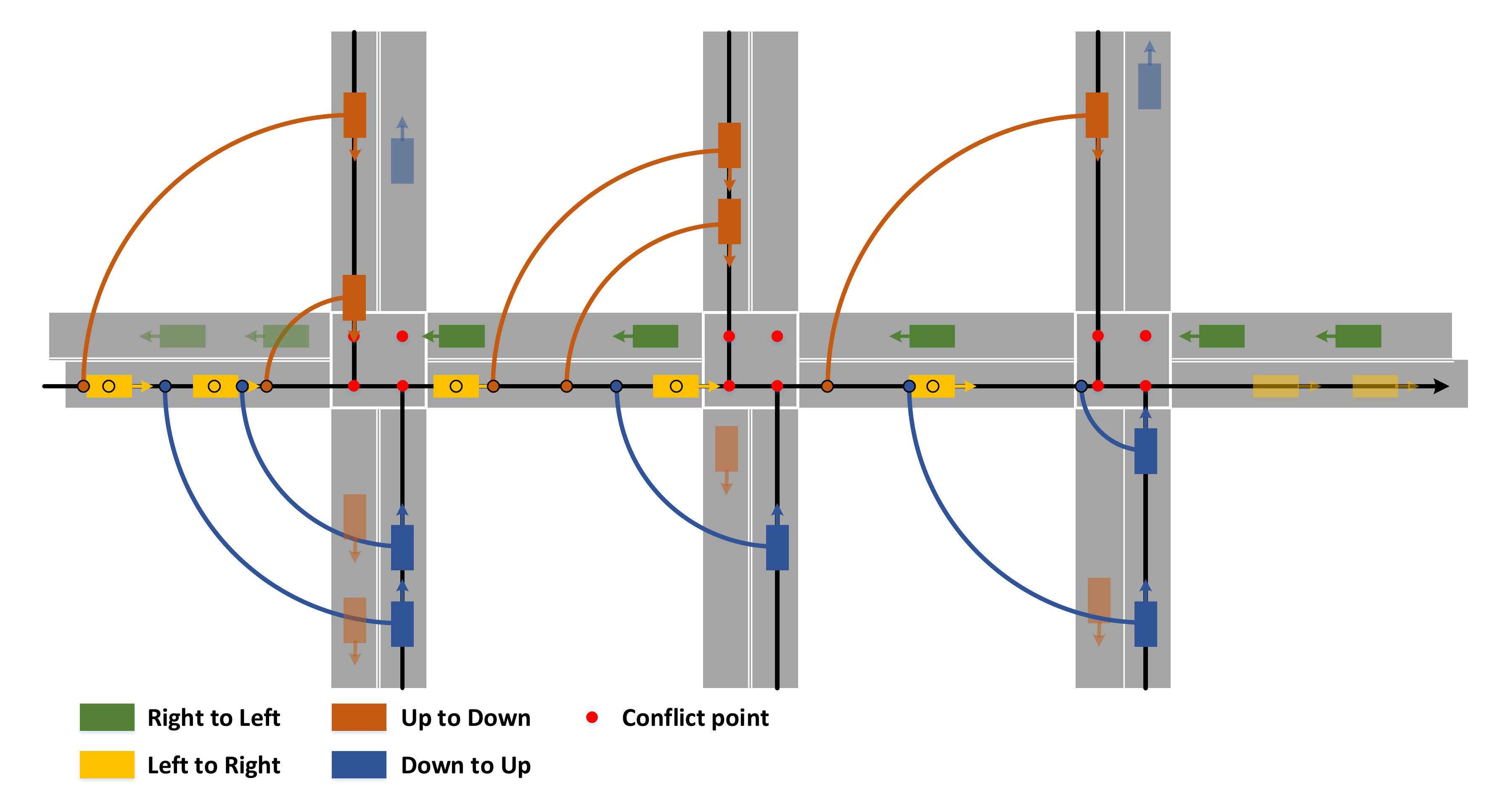}
  \caption{Virtual platoon projection across multiple intersections. Boxes with different colors indicate the vehicles in different directions. Here, yellow vehicles, which runs from left to right, are taken as examples to construct a virtual platoon.}
  \label{fig:virtual_projection} 
\end{figure}

From the perspective of single-vehicle, it is necessary to consider the driving status of reference-vehicles to avoid collisions and improve efficiency. 
This paper introduces the concept of the virtual lane, which is firstly proposed in \cite{xu2018distributed}, as the basis for reference-vehicle selection.

The trajectories of vehicles in different directions will converge into a point, named conflict points. The potential lateral collision occurs at one of the conflict points. 
The conflict vehicles of a specific ego-vehicle are projected onto a base lane with the conflict point as the center. With such operation, the two-dimensional collision avoidance can be converted into one-dimensional collision avoidance.
In the scope of multiple intersections, shown in Fig.\ref{fig:virtual_projection}, the base lane is a long lane across multiple intersections, and conflict vehicles coming from different lanes treat different conflict points as projection centers.
Among the converted one-dimensional vehicles, the nearest vehicles to the ego-vehicle are selected as reference-vehicles, which have the greatest impact on the ego-vehicle.

\subsection{Safety Value}
\label{sec:safety_value}

At a single-intersection area, the gathering of traffic flows results in a smaller inter-vehicle distance, so safety takes priority over efficiency.
This paper defines Safety Value (SV) to describe the vehicle's driving condition from the three aspects: distance, time and acceleration.

The distance SV of vehicle $i$ is calculated as follows,
\begin{equation}
  SV_{i,d}=log((\frac{d_{i,nearest}}{\alpha_{d}})^{\beta_{d}}),
  \label{eq:d_safe_value}
\end{equation}
where $d_{i,nearest}$ denotes the distance between vehicle $i$ and its nearest vehicle on the virtual lane. $\alpha_{d}$ normalizes $d_{i,nearest}$, and it can be treated as the expected headway distance. $\beta_{d}$ increases the offset to improve $log(\cdot)$ effect.

Time safety value $SV_{i,t}$ of vehicle $i$ is calculated as follows,
\begin{equation}
{SV_{i,t}} = \left\{ {\begin{array}{*{20}{c}}
  { - {{\left[ {\frac{{\alpha_{t}}}{{\tanh ( - {t_{i,nearest}})}}} \right]}^{\beta_t}}} & {0 < {t_{i,nearest}} < 1}\vspace{1ex} \\
  2 & {otherwise}
  \end{array}}, \right.
  \label{eq:t_safe_value}
\end{equation}
where $t_{i,nearest}$ denotes Time To Collision (TTC) between vehicle $i$ and its nearest vehicle. In the sensitive range, where $t_{i,nearest}$ is no more than 1, to mark the nearby collision risk, the function $tanh(\cdot)$ is used.

Acceleration safety value $SV_{i,acc}$ of vehicle $i$ is calculated as follows,
\begin{equation}
  \begin{aligned}
&SV_{i,acc} = \\
&\lambda_{acc} \times acc_{i,f} \times \log \left( {\min {{\left( {\frac{{{d_{i,front}}}}{{{d_{threshold}}}},\alpha_{acc}} \right)}^{\beta_{acc}}}} \right),
  \end{aligned}
  \label{eq:acc_safe_value}
\end{equation}
where $d_{i,front}$ is the distance from vehicle $i$ to its front vehicle, $acc_{i,f}$ is the acceleration of the vehicle in front of vehicle $i$, and $d_{threshold}$ is the distance safety threshold. To limit the influence of acceleration in the calculation of safety value, discount factor $\lambda_{acc}$ is introduced.

The combination of SV is calculated as follows,
\begin{equation}
  \begin{aligned}
&{f_{comb}}\left( {{SV_{i,d}},{SV_{i,t}},{SV_{i,acc}}} \right) = \\ 
&clip\left( {\left( {{SV_{i,d}} + {SV_{i,t}} + {SV_{i,acc}}} \right), SV_{max}, SV_{min}} \right),
  \end{aligned}
  \label{eq:safe_value_combination}
\end{equation}
where $SV_{i,d}$, $SV_{i,t}$, and$SV_{i,acc}$ are defined above. In order to suppress the reward explosion, $clip(\cdot)$ is used to limit the maximum and minimum.

With SV, it is easy to evaluate the driving condition of each vehicle numerically. 
The edge nodes select decision-vehicles with the lowest SV.


\subsection{Density}
\label{sec:density}


Closer inter-vehicle will trigger the collision avoidance rules, which causes vehicles to slow down and affects traffic efficiency. Therefore, consciously evenly distributing vehicles is a valuable way to improve traffic efficiency. 
As for vehicle selection, vehicles with too sparse or dense distribution should be selected as decision-vehicles. The evaluation of the density is based on the average density of the entire lane. We define the criterion with Density Indicator ($DI_i$) for each vehicle, which as follows,
\begin{equation}
  D{I_i} = \left| {\frac{{\left| {{p_{f,i}} - {p_{l,i}}} \right|}}{2n} - \frac{{\left| {{p_f} - {p_l}} \right|}}{{{N_L} - 1}}} \right|,
  \label{eq:density_indicator}
\end{equation}
where $p_{f,i}$, is the position of the $n^{th}$ vehicle before vehicle $i$, and $p_{l,i}$ is the position of the $n^{th}$ vehicle after vehicle $i$.
$\frac{{\left| {{p_f} - {p_l}} \right|}}{{{N_L} - 1}}$ indicates the average inter-vehicle distance on the entire lane. With the operation of $\left| \cdot \right|$, vehicles with too sparse or dense distribution can be selected by the cloud node as decision-vehicles, which have the highest $DI_i$.

\section{The Architecture of MiVeCC}

\subsection{General}

\begin{figure}
  \centering
  \includegraphics[width=\linewidth]{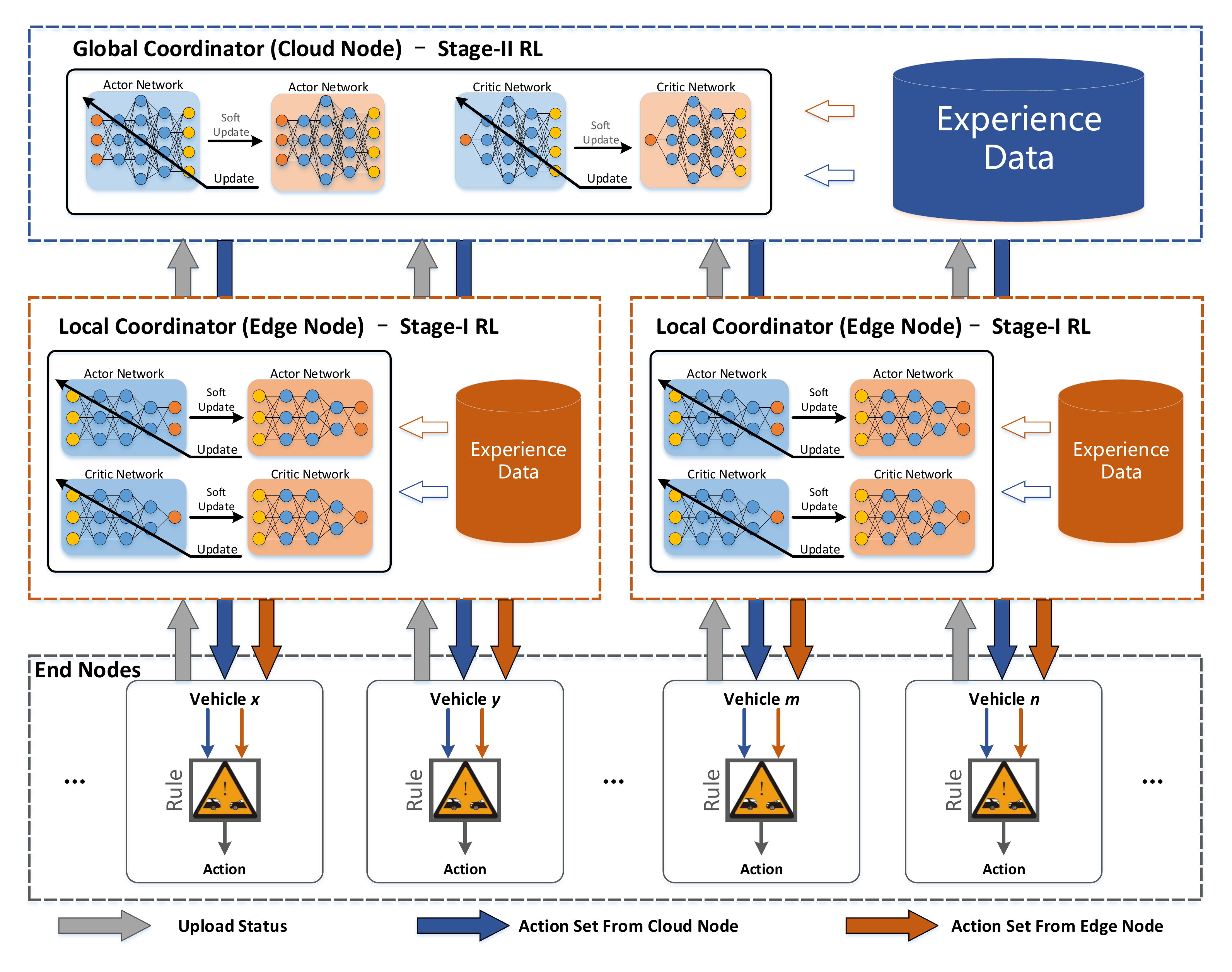}
  \caption{The overall architecture of MiVeCC. The end nodes operate in the environment and upload status to the local and global coordinator. The coordinators produce actions for end nodes. Meanwhile, the coordinators evaluate the status and update the NNs for decision-making.}
  \label{fig:alg_arch} 
\end{figure}

This part illustrates how to solve vehicle cooperation in a large area with multiple intersections in details.

The proposed MiVeCC consists of three parts: cloud node (global coordinator), edge nodes (local coordinator), and end nodes (vehicle executor).
Fig.\ref{fig:alg_arch} shows the overall architecture of the scheme. The end nodes interact with the environment and upload the driving status (position, velocity, and acceleration) to the edge nodes and the cloud node.
The edge nodes produce the action set for vehicles in their scope, and the cloud node produces the action set for all vehicles in the large area. In the above process, cloud and edge nodes can evaluate the actions under the specific condition and generate the experience data to support the training process of RL.

In this paper, the RL algorithm adopted is a variant based on Deep Deterministic Policy Gradient (DDPG)\cite{DBLP:journals/corr/LillicrapHPHETS15}, which is good at continuous action control.

\subsection{Reinforcement Learning Training}

The two stages take DDPG as the RL algorithm. 
Edge nodes perform stage-I RL, and the cloud nodes perform stage-II RL.
An RL agent needs to obtain experience for model training by continuously interacting with the environment before it can reason ideally.

Firstly, the agent obtains action with
\begin{equation}
  {a_t} = \mu ({s_t}|{\theta ^\mu }) + {\mathcal{N}_t},
\end{equation}
where $\mu (\cdot|{\theta ^\mu })$ is a parameterized actor module in the evaluation network, and ${\mathcal{N}_t}$ is noise for action exploration.
Secondly, the agent performs the action $a_t$ to obtain experience $(s_t,a_t,r_t,s_{t+1})$, which contains the current state, action, reward and next state, and stores it in the replay buffer.
Thirdly, the agent samples a random minibatch from the replay buffer to update actor module and critic module.
The agent updates critic by minimizing the loss $L$ which are as follows,
\begin{equation}
  \begin{array}{c}
    {y_i} = {r_i} + \gamma Q'({s_{i + 1}},\mu '({s_{i + 1}}|{\theta ^{\mu '}})|{\theta ^{Q'}})\\
    L = \frac{1}{B}\sum\nolimits_i {{{({y_i} - Q({s_i},{a_i}|{\theta ^Q}))}^2}} 
    \end{array},
\end{equation}
where $\mu' (\cdot|{\theta ^{\mu'} })$ and $Q' (\cdot|{\theta ^{Q'} })$ are actor module and critic module in the target network, $Q (\cdot|{\theta ^Q })$ is parameterized critic module in the evaluation network. 
Discounted factor $\gamma$ is used to control the effect of future reward on the present.
$B$ is the size of minibatch, which is the number of experience extracted from replay memory for training.
The agent updates actor policy using the sampled policy gradient, which is as follows,
\begin{equation}
  {\nabla _{{\theta ^\mu }}}J = \frac{1}{B}\sum\nolimits_i {{\nabla _a}Q(s,a|{\theta ^Q}){\nabla _{{\theta ^\mu }}}\mu (s|{\theta ^\mu })}.
\end{equation}

Finally, the parameters of actor and critic in target network ($\theta ^Q, \theta ^\mu$) are updated by slowly tracking the evaluation network ($\theta ^{Q'},\theta ^{\mu'}$), which are as follows,
\begin{equation}
  \begin{array}{l}
    {\theta ^{Q'}} \leftarrow \tau {\theta ^Q} + (1 - \tau ){\theta ^{Q'}}\\
    {\theta ^{\mu '}} \leftarrow \tau {\theta ^\mu } + (1 - \tau ){\theta ^{\mu '}}
    \end{array}.
\end{equation}

\subsection{Cloud Node}

The cloud node collects the vehicle information in the whole system to form the vehicle set $V_{cloud}$. After evaluating each vehicle's density on each virtual lane across multiple intersections, the cloud node selects the vehicles with the highest $DI$ to generate State Graph $SG_{cloud,i}$ for each virtual lane $VL_i$. With $SG_{cloud,i}$, the cloud node can get action set $A_{cloud,i}$ with NN. The entire process is presented in Algorithm \ref{alg:cloud_decision}, where $f_{NN,cloud}(\cdot)$ means the inference process with NN on the cloud.

\begin{algorithm}
  \caption{Decision-Making on The Cloud Node}
  \LinesNumbered
  \KwIn{
    \\$\hspace{2ex}$Vehicle set $V_{cloud}$
    }
  \KwOut{
    \\$\hspace{2ex}$Vehicle action set $A_{cloud}$
    }
  \label{alg:cloud_decision}

  Construct the virtual lane set $VL$ for each actual lane\;
  \ForEach{$VL_i$ in $VL$}{
    \ForEach{actual vehicle  on $VL_i$}{
      Calculate the $DI_{i,j}$ with eq.(\ref{eq:density_indicator}) for $v_{i,j}$\;
    }
    Select the decision-vehicle set $DV_i$ with the highest $DI$\;
    Select the reference-vehicle set $RV_i$\ with the distance on $VL_i$\;
    Construct $SG_{cloud,i}$ with $DV_i$ and $RV_i$\;
    $A_{cloud,i} = f_{NN,cloud}(SG_{cloud,i})$\;
  }
  Combine each $A_{cloud,i}$ to get $A_{cloud}$.
  
\end{algorithm}

\subsection{Edge Node}

The edge node collects the vehicle information in its scope to form a vehicle set $V_{edge}$. After calculating each vehicle's $SV_i$ in the set, the edge node selects the vehicles with the lowest $SV_i$ to generate State Graph $SG_{edge}$. With $SG_{edge}$, the edge node can obtain action $A_{edge}$ with the NN. The entire process is presented in Algorithm \ref{alg:edge_decision}, where $f_{NN,edge}(\cdot)$ means the inference process with NN on the edge nodes.

\begin{algorithm}
  \caption{Decision-Making on Each Edge Node}
  \LinesNumbered
  \KwIn{
    \\$\hspace{2ex}$Vehicle set $V_{edge}$
    }
  \KwOut{
    \\$\hspace{2ex}$Vehicle action set $A_{edge}$
    }
  \label{alg:edge_decision}

  Construct the virtual lane set $VL$ for each actual lane at the single intersection\;

  \ForEach{actual vehicle $v_{i}$ on $VL$}{
    Calculate safety value $SV_{i}$ for $v_{i}$ with Eq.(\ref{eq:safe_value_combination})\;
  }

  Select the decision-vehicle set $DV$ with the lowest $SV_{i}$\;
  Select the reference-vehicle set $RV$\ with the distance on the corresponding virtual lane\;
  Construct the State Graph $SG_{edge}$ with $DV$ and $RV$\;
  $A_{edge} = f_{NN,edge}(SG_{edge})$\;  
  
\end{algorithm}

\subsection{End Node}

RL relies on the balance between exploration and exploitation to obtain the optimal strategy.
However, the uncertainty of exploration seriously hinders the acquisition of the optimal strategy, especially in the traffic environment, which will bring more probabilities of collision. Although the rule-based vehicle control scheme can entirely guarantee safety, it is challenging to design the best performance strategy.

To achieve better performance in vehicle cooperative control at unsignalized intersections, the vehicle behavior is decided by the combination of end, edge and cloud nodes. As presented in Algorithm \ref{alg:onboard_decision}, each vehicle can receive the cloud action instruction $a_{cloud}$ and edge action instruction $a_{edge}$. Each vehicle calculates onboard action $a_{end}$, and obtains the action to be executed $a_{exe}$ based on $a_{end}$, $a_{edge}$, and $a_{cloud}$.


\begin{algorithm}
  \caption{Decision-Making on Each End Node}
  \LinesNumbered
  \KwIn{
    \\{$\hspace{2ex}$Action from the cloud node $a_{cloud}$}
    \\{$\hspace{2ex}$Action from edge node $a_{edge}$}
    \\{$\hspace{2ex}$Distance to vehicle in front $d_f$}
    \\{$\hspace{2ex}$Distance to vehicle behind $d_b$}
    }
  \KwOut{
    \\$\hspace{2ex}$Action to be executed $a_{exe}$
    }
  \label{alg:onboard_decision}
    
  Construct the virtual lane based on the ego-vehicle\;
  Calculate $SV$ for the ego-vehicle with Eq.(\ref{eq:safe_value_combination})\;
  Calculate $a_{end}$ with
  \begin{equation*}
    \left\{ {\begin{array}{*{20}{c}}
      {\left| {\frac{{SV}}{\eta}} \right|}&{d_f \leq d_b}\vspace{1ex} \\
      {\frac{{SV}}{\eta}}&{d_f \textgreater d_b}
      \end{array}} \right.;
    \end{equation*}

  $a_{exe}=a_{end}$\;

  \If {$a_{edge}$ exists}{
    Calculate $a_{end+edge}$ with
    \begin{equation*}
      \begin{array}{l}
        \left\{ {\begin{array}{*{20}{c}}
        {\max ({a_{edge}},{a_{end}})}&{{a_{edge}} \times {a_{end}} > 0}\vspace{1ex}\\
        {{a_{end}}}&{otherwise}
        \end{array}} \right.
        \end{array};
      \end{equation*}

    $a_{exe}=a_{end+edge}$\;
  }

  \If {$a_{cloud}$ exists}{
    $a_{exe} = (1-\omega) \times a_{exe} + \omega \times a_{cloud}$\;
  }

\end{algorithm}

\section{Simulation and Results}

In this paper, a newly designed simulation platform is developed based on Python 3.5. The platform aims to verify the performance of vehicle control in both the single-intersection and multi-intersection scenarios without traffic lights.
The generation of traffic flows obeys Poisson distribution, and the traffic density is controllable. Both centralized and decentralized schemes can be examined in the platform. 
Besides, the end-edge-cloud structure is also a key point for the platform. The nodes in the three layers (end layer, edge layer and cloud layer) can communicate with each other.
This platform takes into the control range account at multiple intersections, vehicles can be controlled by the corresponding edge nodes in sequence. The simulation platform is released at: XXXXXXXXXXXXXXXXXXXXX.

The entire experiment has three parts.
Firstly, to verify the effectiveness of the proposed end-edge-cloud computing architecture, this paper demonstrates the average changing velocity of vehicles changes with the RL model's training process. Secondly, in order to verify the high performance of the proposed algorithm, this paper compares the MiVeCC with other schemes in terms of traffic efficiency. Finally, to further illustrate that the unsignalized solution  is better than the solution with traffic lights, the heat map is used to show the instantaneous vehicle velocity and vehicle density distribution under the multi-intersection scenario with the highest density (2100 vehicles/hour/lane).

\subsection{Experimental Settings}

In this part, our proposed scheme is trained and evaluated at the $3\times3$ multi-intersection scenario, which consisting of 4 driving directions, and vehicles are allowed to go straight without steering. Each intersection contains 4 conflict points and 4 virtual lanes. Besides, there are 4 types of vehicles from different directions, following the Poisson distribution with different densities. The related parameters are listed in the table \ref {tab:Experiment_Parameter}. Each intersection is equipped with an edge node, collecting vehicle kinetic information, and then guiding vehicles hold in this range after training the edge-RL model. Unlike edge nodes, the cloud node is set up to collect vehicle kinetic information and guide the vehicles in the whole multi-intersection scenario after training the cloud-RL model.

\begin{table}
  \centering
  \caption{Experimental Parameters}
  \label{tab:Experiment_Parameter}
  \begin{tabular}[l]{@{}lc}
  \toprule
      \textbf{Parameter} & \textbf{Value} \\
      \midrule
      
      \textbf{\emph{Simulator}} \\
      Lane length ($m$) & 150 \\ 
      Vehicle size ($m$) & 2 \\ 
      Velocity ($m/s$) & [6,13] \\ 
      Initial velocity ($m/s$) & 10 \\ 
      Acceleration ($m/s^2$) & [-3,3] \\ 
      Discrete-time step $T$ ($s$) & 0.1 \\ 
      
      \midrule
      \textbf{\emph{Safety Value}} \\
      $\alpha_{d}$ & 10 \\ 
      $\beta_{d}$ & 10 \\ 
      $\alpha_{t}$ & 1.5 \\ 
      $\beta_{t}$ & 2 \\ 
      $\alpha_{acc}$ & 1.5 \\ 
      $\beta_{acc}$ & 12 \\ 
      $\lambda_{acc}$ & 0.2 \\ 
      $SV_{max}$ & 20 \\ 
      $SV_{min}$ & -20 \\ 
      
      Conversion factor $\eta$ & 3 \\
      Fusion factor $\omega$ & 0.2 \\

      \midrule
      \textbf{\emph{Vehicle Selection}} \\
      Number of the closet vehicle $n$ & 5 \\

    \bottomrule
  \end{tabular}
\end{table}

\subsection{The Details of Neural Network}

\begin{figure}
  \centering
  \includegraphics[width=\linewidth]{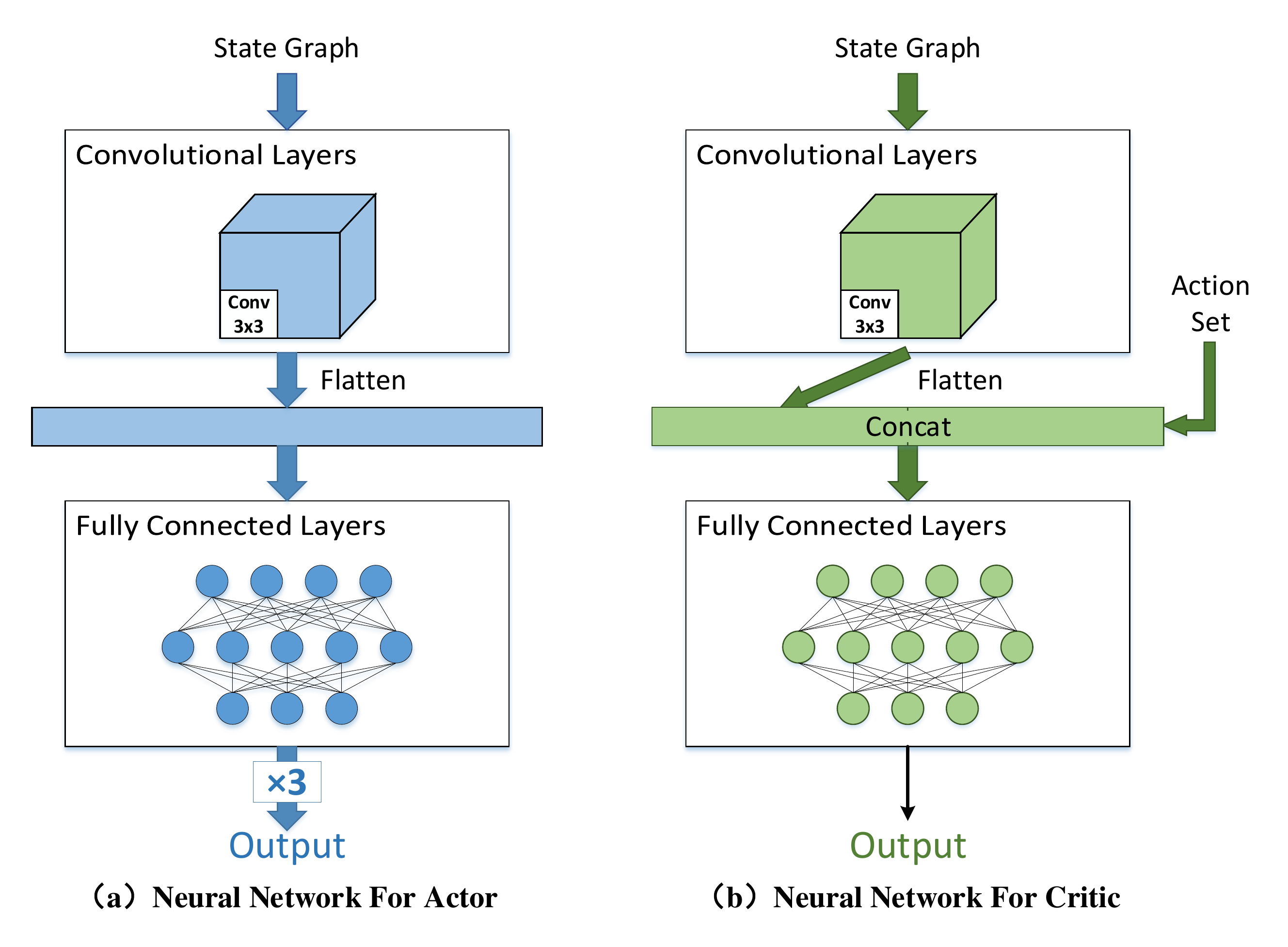}
  \caption{The neural network for the actor (a) and the critic (b). Actor and critic have the similar network structure. The differences between the two is that the output of the actor is multiplied by 3, and that the action set is concatenated to the intermediate value produced by the convolution layers.}
  \label{fig:NN} 
\end{figure}

In this paper, NNs are deployed as an approximate function of actor and critic on cloud and edge nodes, including convolutional layers and fully connected layers. The details are shown in Fig.\ref{fig:NN}.

On the cloud node, the SG input is in the form of a three-dimension array (60, 60, 3), where each element indicates the number of rows, columns and channel. For both actor and critic, there are three convolutional layers. The output of the convolutional layers is flattened into a one-dimensional vector, then as the input of dense layers. In the actor part, to fit the value range of action, the dense layer's output array is multiplied by 3 times. The critic part is used to evaluate the action set under the specific SG. On the edge node, the input is also a three-dimension array (15, 15, 3). 
The edge node has a similar network structure with the cloud node in both actor and critic.
The complete hyperparameters are listed in TABLE \ref{tab:HyperParameter}. For actor, $tanh(\cdot)$ is adopted as activation function. For critic, ReLU is adopted as activation function, and layer normalization is also adopted.

\begin{table}
  \centering
  \caption{Parameters for Neural Networks}
  \label{tab:HyperParameter}
  \begin{tabular}[l]{@{}lccccccc}
  \toprule
      \makecell[c]{\textbf{Parameter}} & \multicolumn{4}{c}{\textbf{Value}}\\ 
      
      \midrule
      Discounted factor $\gamma$ & \multicolumn{4}{c}{0.8}\\
      Minibatch $B$ & \multicolumn{4}{c}{48} \\ 
      Soft update factor $\tau$ & \multicolumn{4}{c}{0.99} \\ 
      Episode & \multicolumn{4}{c}{50} \\ 
      Learning rate-actor & \multicolumn{4}{c}{0.001 $\rightarrow$ 0} \\ 
      Learning rate-critic & \multicolumn{4}{c}{0.05 $\rightarrow$ 0} \\ 
      Optimizer & \multicolumn{4}{c}{Adam} \\ 

      \midrule
      \makecell[c]{\textbf{Parameter}} & \multicolumn{2}{c}{\textbf{Cloud}} & \multicolumn{2}{c}{\textbf{Edge}}\\
      \midrule
      
      State graph input& \multicolumn{2}{c}{(60,60,3)} & \multicolumn{2}{c}{(15,15,3)} \\
      Action set input& \multicolumn{2}{c}{60} & \multicolumn{2}{c}{15} \\
      Convolutional layer 1\# & \multicolumn{2}{c}{(32, 3$\times$3, 2$\times$2, valid)} & \multicolumn{2}{c}{(32, 3$\times$3, 2$\times$2, valid)} \\
      Convolutional layer 2\# & \multicolumn{2}{c}{(32, 3$\times$3, 2$\times$2, valid)} & \multicolumn{2}{c}{(16, 3$\times$3, 2$\times$2, valid)} \\
      Convolutional layer 3\# & \multicolumn{2}{c}{(16, 3$\times$3, 2$\times$2, valid)} & \multicolumn{2}{c}{-} \\
      Flattened elements & \multicolumn{2}{c}{576} & \multicolumn{2}{c}{144} \\

      \midrule
      \makecell[c]{\textbf{Parameter}} & \textbf{Actor} & \textbf{Critic} & \textbf{Actor} & \textbf{Critic} \\
      \midrule
      Dense layer 1\# & 512 & 512 & 96 & 96\\
      Dense layer 2\# & 256 & 256 & 96 & 96\\
      Dense layer 3\# & 60 & 128 & 15 & 48\\
      Dense layer 4\# & - & 54 & - & 12 \\
      Dense layer 5\# & - & 1 & - & 1 \\
    \bottomrule
  \end{tabular}
  \footnotesize{Note: Each element in the convolutional layer four-tuple indicates the filter number, filter size, stride size and padding type separately.}
\end{table}

\subsection{Simulation Results}

\begin{figure}
  \centering
  \includegraphics[width=\linewidth]{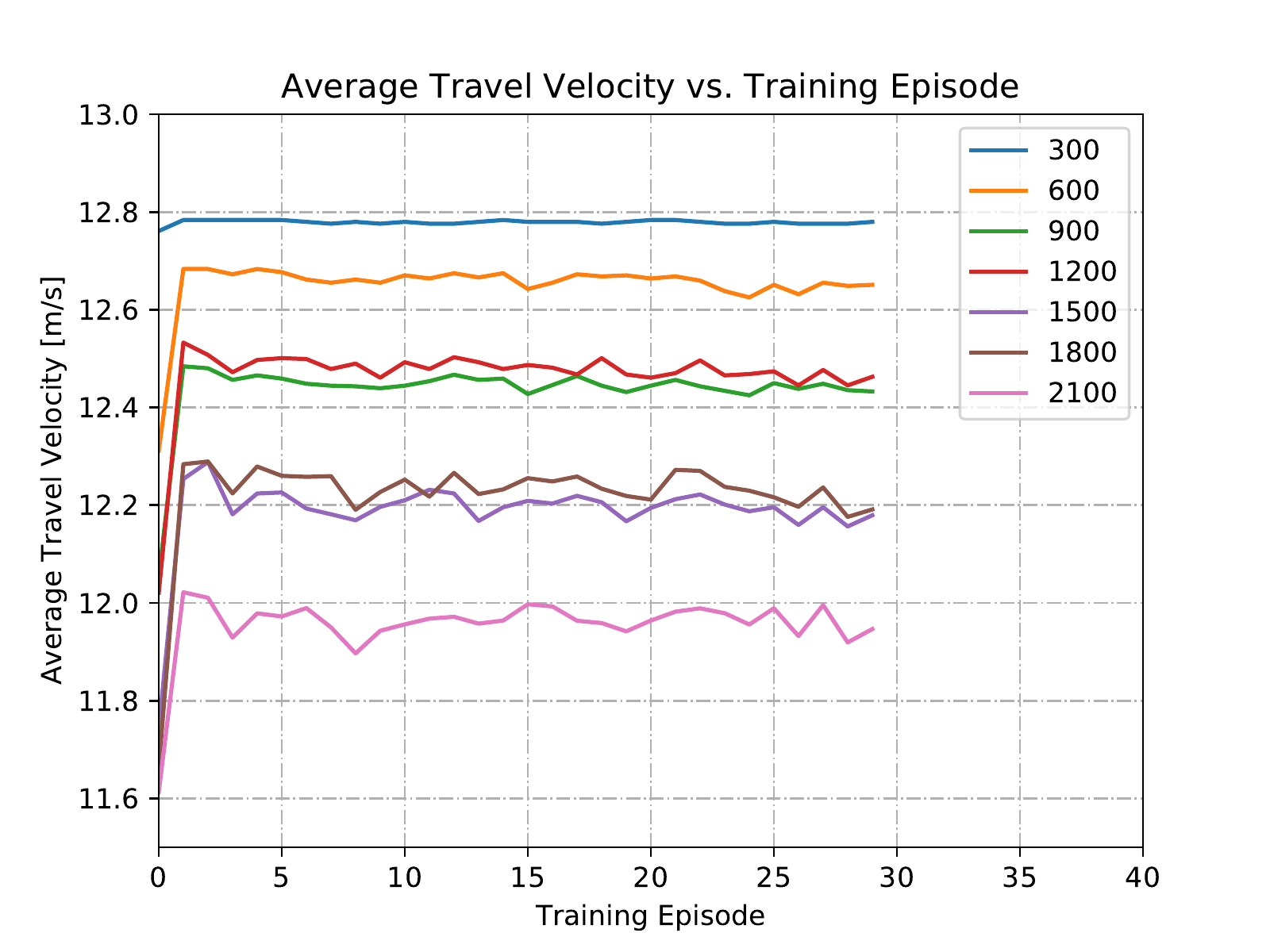}
  \caption{Performance of policies with the training episodes under different vehicle densities at a single intersection.}
  \label{fig:single_train} 
\end{figure}

In the first experiment, a policy trained by our proposed scheme is evaluated at single-intersection. Fig.\ref{fig:single_train} shows the relationship between the average travel velocity and the training episode under the scenarios of different densities. 
Different color lines represent the results of average velocities under different densities scenarios, respectively. 
The blue and pink solid lines represent the conditions when the number of vehicles is 300 vehicles/lane/hour (low density) and 2100 vehicles/lane/hour (high density), respectively. 
As is shown in Fig.\ref{fig:single_train}, with the training episode increasing, the policy quickly converges to stability under each density scenario, which only runs a few episodes. 

\begin{figure}
  \centering
  \includegraphics[width=\linewidth]{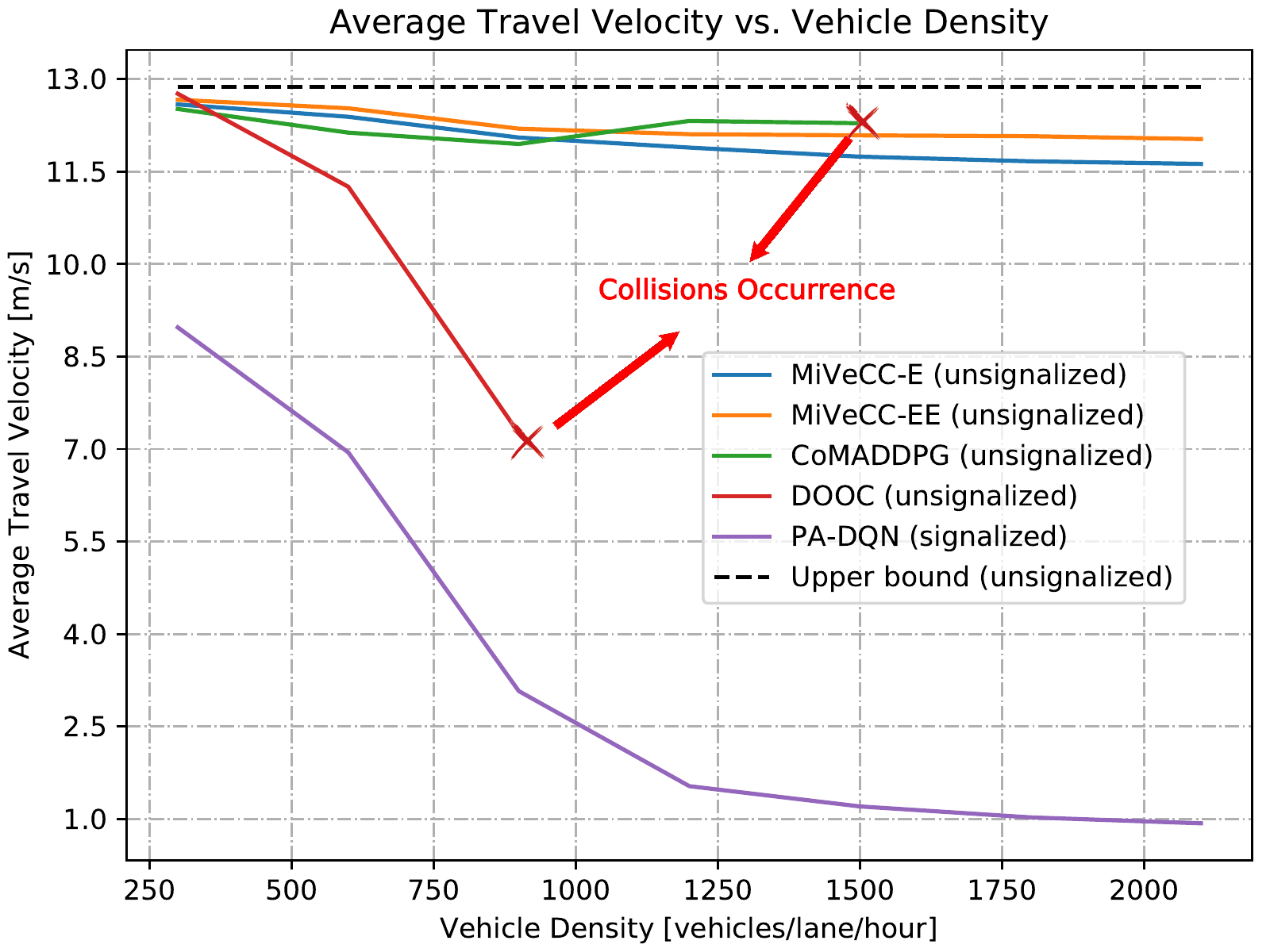}
  \caption{Comparison of multiple schemes at a single intersection.}
  \label{fig:single_compare} 
\end{figure}

Fig.\ref{fig:single_compare} compares the performances between MiVeCC and several benchmark schemes.
The blue line, named MiVeCC-E, denotes the proposed collsion avoidance rules operated by vehicles, and the orange line, named MiVeCC-EE, represents the proposed scheme which combines the rules and the local decision of the edge node. 
The purple line illustrates an application of DQN, named PA-DQN, in the single intersection traffic light control, mentioned in \cite{DBLP:journals/tcyb/TanBDJDW20}. The red line shows an optimization scheme, named DOOC, proposed in \cite{DBLP:journals/tie/BianLRWLL20}, and the green line presents a distributed RL scheme, named CoMADDPG, proposed in \cite{wu2020cooperative}. 
The black dotted line denotes the average velocity of a vehicle passing the intersection at the fastest, which is the theoretical upper bound. 
As is shown in this figure, the signalized scheme can guarantee traffic safety, but the performance gets worse significantly. What's more, DOOC produces collisions when the density exceeds 900 vehicles/lane/hour.
Also, CoMADDPG attempts to obtain better policy than the proposed scheme, but collisions occur when the density exceeds 1500 vehicles/lane/hour.
As a result, the combination of the proposed schemes performs better than rules only all the time.

\begin{figure}
  \centering
  \includegraphics[width=\linewidth]{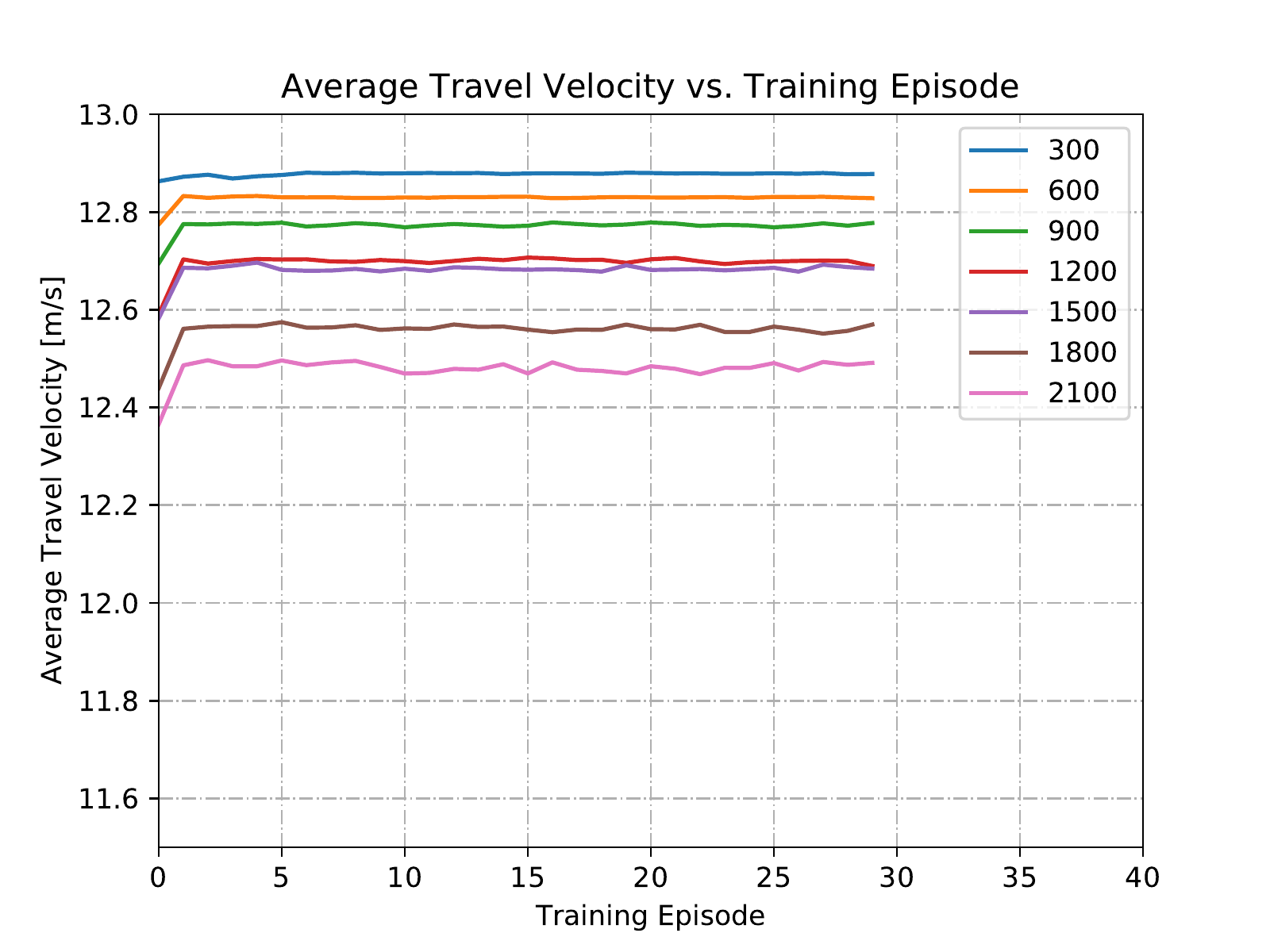}
  \caption{Performance of policies with the training episodes under different vehicle densities at multiple intersections. 
  }
  \label{fig:multiple_train} 
\end{figure}

In the second experiment, the policy trained by the proposed scheme is evaluated at multiple intersections.
Fig.\ref{fig:multiple_train} further researches the policy with the trained edge model under different vehicle densities. Firstly, as the vehicles get denser, the average velocity of vehicles decreases. More importantly, average velocities combining the trained edge model start at a relatively high value under different densities. 

\begin{table*}
  \centering
  \caption{Efficiency Improvement Rate}
  \label{tab:eff_imp_rate}
  \begin{tabular}{|c|c|c|c|c|c|c|c|}
  \hline
       \multirow{2}{*}{\textbf{Schemes}} &  \multicolumn{7}{c|}{\textbf{Vehicle Density [vehicles/lane/hour]}}  \\ \cline{2-8}
       & 300 & 600 & 900 & 1200 & 1500 & 1800 & 2100 \\ \hline
      MiVeCC-EEC vs. Coder & 79.80\% & 85.10\% & 146.63\% & 283.32\% & 352.26\% & 419.48\% & 459.09\% \\ \hline
      MiVeCC-EE vs. MiVeCC-E & 0.14\% & 0.40\% & 0.52\% & 0.91\% & 1.47\% & 1.92\% & 2.12\% \\ \hline
      MiVeCC-EEC vs. MiVeCC-E & 0.37\% & 1.00\% & 1.56\% & 1.96\% & 2.32\% & 2.76\% & 2.96\% \\ \hline
      MiVeCC-EEC vs. MiVeCC-EE & 0.23\% & 0.6\% & 1.03\% & 1.04\% & 0.84\% & 0.83\% & 0.82\% \\ \hline
  \end{tabular}
\end{table*}

\begin{figure}
  \centering
  \includegraphics[width=\linewidth]{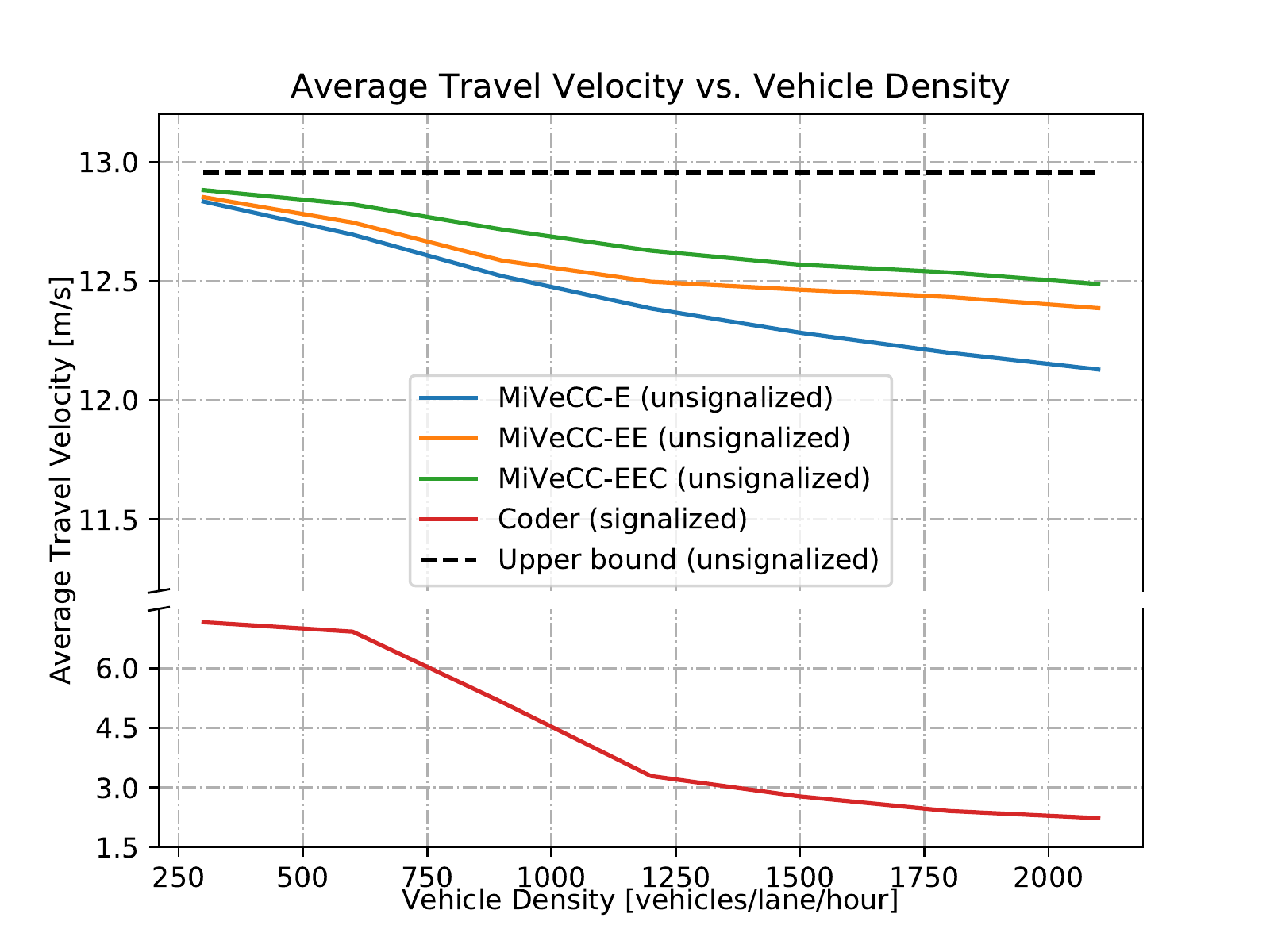}
  \caption{Comparison of multiple schemes at multiple intersections.}
  \label{fig:multiple_compare} 
\end{figure}

Fig.\ref{fig:multiple_compare} presents the performance comparison of the proposed MiVeCC and one benchmark scheme.
MiVeCC-EEC denotes the results of the proposed scheme which combines three aspects, including rules, the local decision of edge nodes and the global decision of cloud node.
Coder, proposed in \cite{DBLP:journals/tcyb/TanBDJDW20}, is a cooperative deep reinforcement learning framework applied in signalized intersection control to achieve cooperation among intersections.
The black dotted line denotes the theoretical upper bound as Fig.\ref{fig:single_compare}.
The results show that the proposed unsignalized scheme performs better than the signalized scheme at multiple intersections. 
As shown in TABLE \ref{tab:eff_imp_rate}, the performance improvement with the proposed unsignalized scheme continues to expand as the density increases until the highest density, where the improvement rate reaches 4.59 times.
Although signal-Coder can handle multiple signalized intersection control, the performance drops seriously as density increases.
What's more, with the superimposition of the number of vertical cooperation layers, the traffic efficiency is also correspondingly improved. Still, when the velocities are close to the theoretical upper bound, the room for improvement becomes marginal.

\begin{figure}
  \centering
  \subfigure[MiVeCC]{\includegraphics[width=0.45\linewidth]{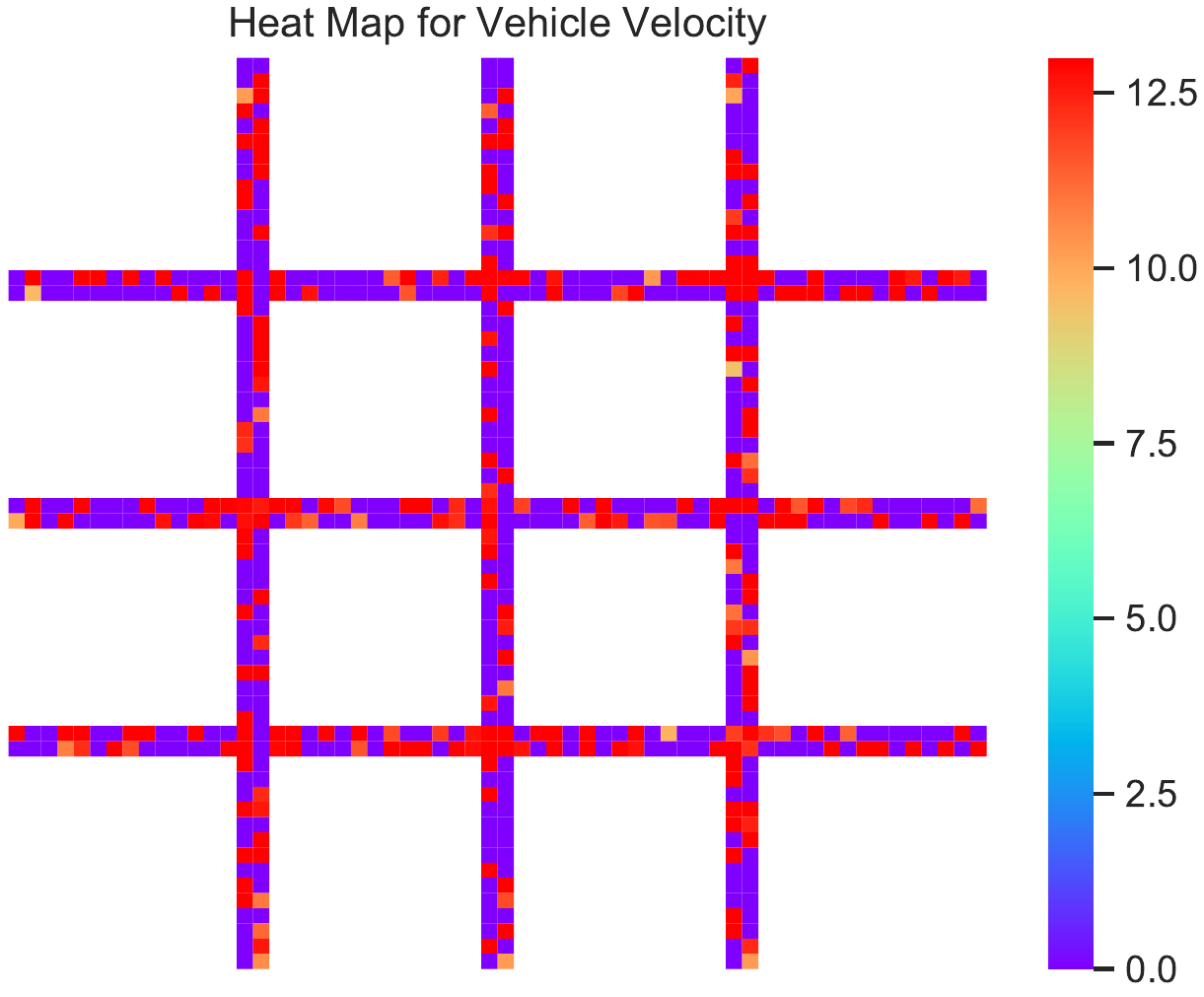}}
  \subfigure[Coder]{\includegraphics[width=0.45\linewidth]{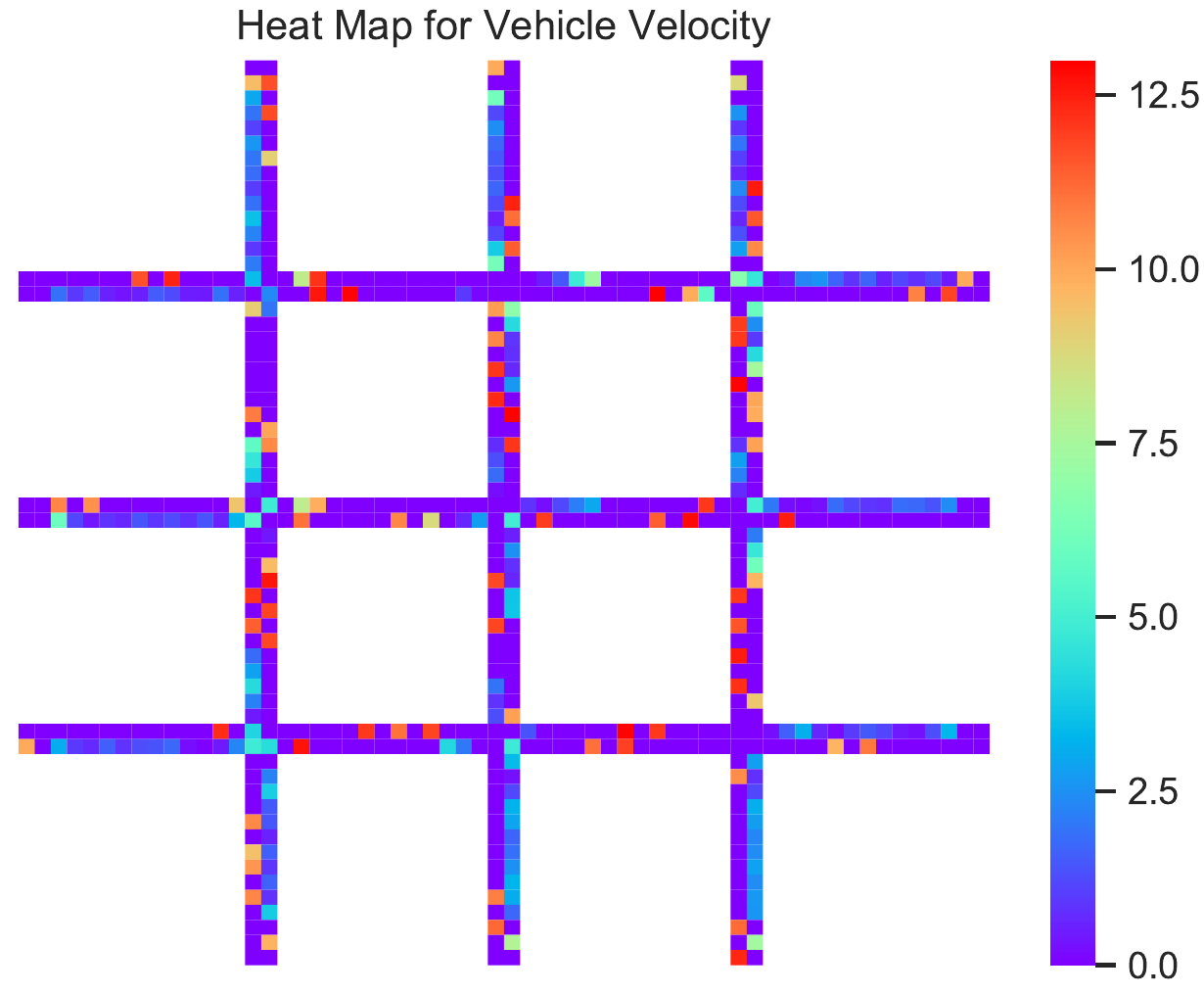}}
  \caption{The comparison of the velocity heat maps between the proposed scheme (MiVeCC) and the benchmark scheme (Coder).}
  \label{fig:heat_velocity}
\end{figure}

Fig.\ref{fig:heat_velocity} are heat maps of vehicle velocity distribution with the proposed scheme and the benchmark scheme, which is Coder mentioned above.
The data in is sampled at a certain time step during the evaluation under the density of 2100 vehicles/lane/hour at multiple intersections.
The purple grid indicates that there is no vehicle inside. The red grid represents that the vehicle driving at a high velocity, and grids with other colors mean that the vehicle driving at a low velocity.
It can be observed that the proposed scheme can significantly increase the velocity.
Besides, with the proposed scheme, most vehicles are close to the maximum velocity, indicating that the vehicle velocity fluctuation is small.
Therefore, the proposed scheme is more efficient than the benchmark scheme.

\begin{figure}
  \centering
  \subfigure[MiVeCC]{\includegraphics[width=0.45\linewidth]{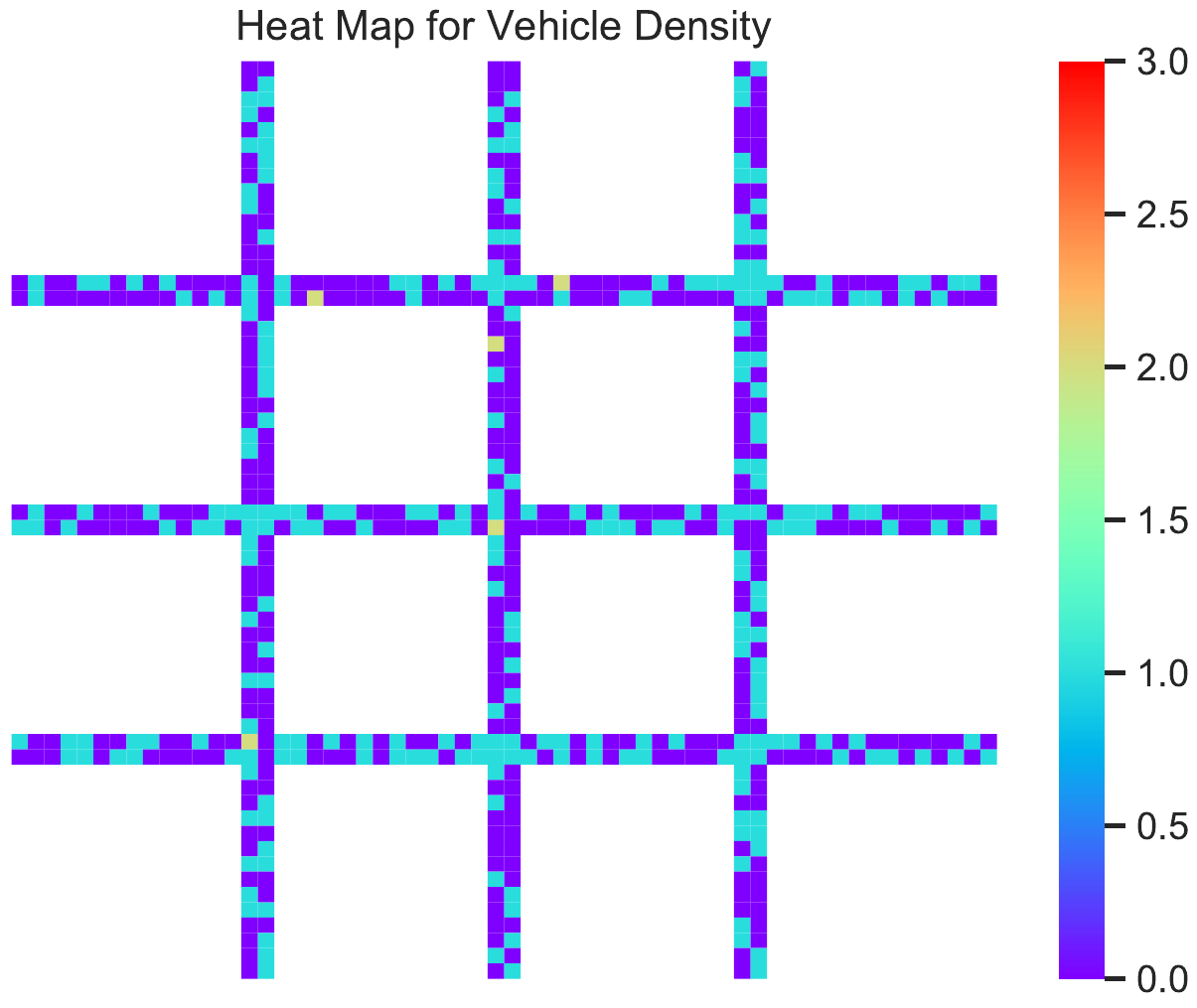}}
  \subfigure[Coder]{\includegraphics[width=0.45\linewidth]{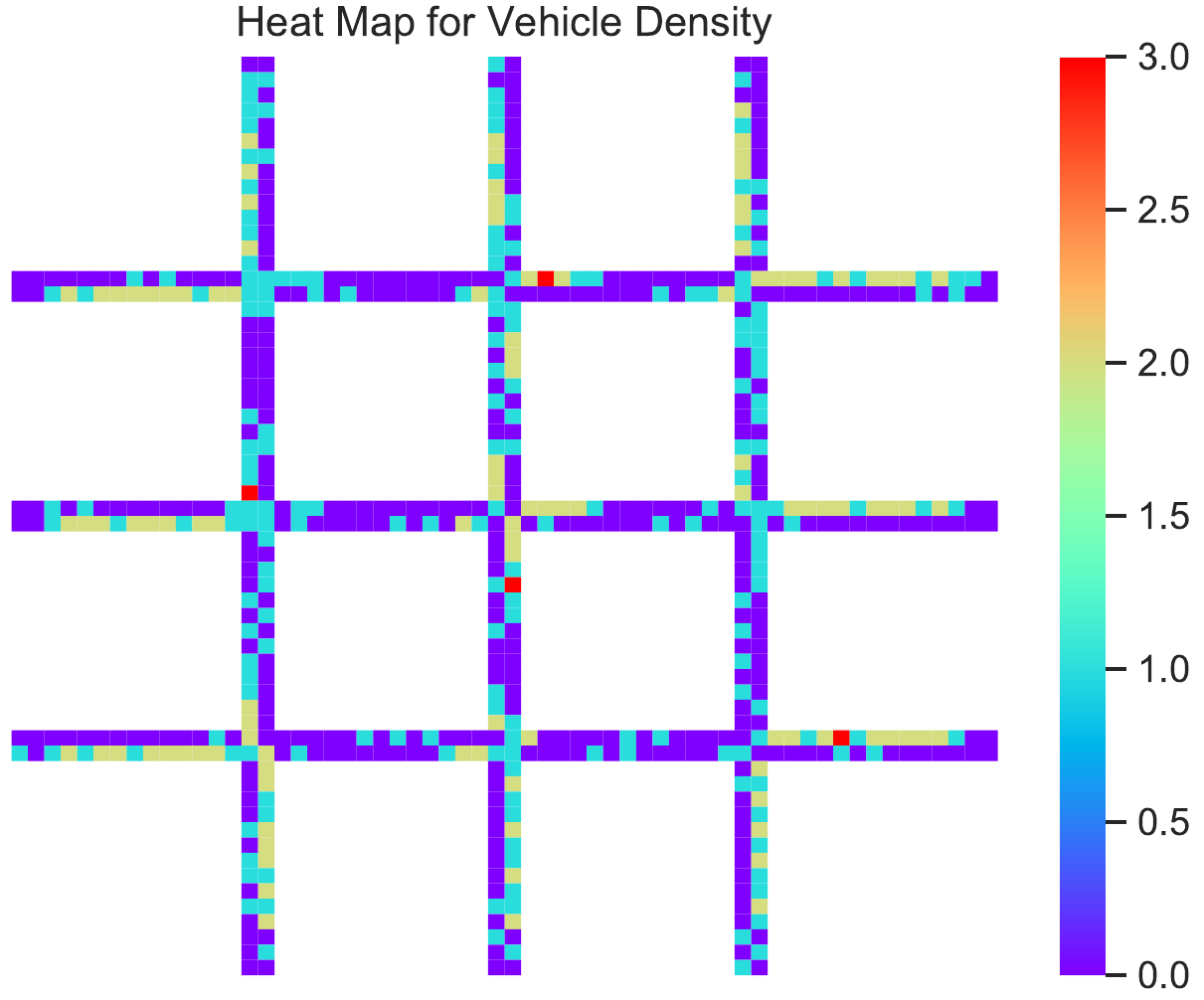}}
  \caption{The comparison of the density heat maps between the proposed scheme (MiVeCC) and the benchmark scheme (Coder).}
  \label{fig:heat_density}
\end{figure}

Fig.\ref{fig:heat_density} are the heat maps of spatial distribution of vehicle for the proposed unsignalized scheme and the benchmark scheme under the density of 2100 vehicles/hour/lane, where the sampling time of traffic flow density is consistent with Fig.\ref{fig:heat_velocity}.
It can be seen that, the proposed unsignalized scheme contributes to vehicles more evenly distributed at intersections without congestion.
As a result, the proposed unsignalized scheme can improve the utilization rate of the road, and has significant advantages in solving urban congestion, improving traffic efficiency.

\section{Conclusion}

In this paper, we propose a new vehicle control scheme, namely the MiVeCC, which contains a vehicular end-edge-cloud computing framework, two-stage reinforcement learning and vehicle selection methods.
With the aggregation of end-edge-cloud vertical cooperation and horizontal cooperation among vehicles, MiVeCC can find a well-performed vehicular cooperative control policy in a large area with multiple unsignalized intersections.
Incorporating the idea of vehicle selection methods, the state space can be reduced to accelerate the algorithm convergence without performance degradation.
A new simulation platform is developed to verify the proposed MiVeCC's performance at different intersections. The simulation results show that MiVeCC has outstanding performance even in the high-density scenario, which improves traffic efficiency up to 4.59 times compared with existing methods. 
To the best of our knowledge, this is the first attempt to address the vehicular cooperative control problems based on end-edge-cloud computing. We believe that this paper is a well-performing alternative to the existing schemes of autonomous driving at intersections.

Our main future work will be focused on tackling further optimization in vehicle control at multiple unsignalized intersections. Platoon refers to an organization in which multiple vehicles form a closed group. Although the research on platoon exists for years, the application of platoon at intersections is extremely attractive, especially the dynamic change of platoon.

\section*{Acknowledgment}

The authors would like to thank...

\ifCLASSOPTIONcaptionsoff
  \newpage
\fi



%

\bibliographystyle{IEEEtran}
\bibliography{IEEEabrv,ref}




%

\begin{IEEEbiography}[{\includegraphics[width=1in,height=1.25in,clip,keepaspectratio]{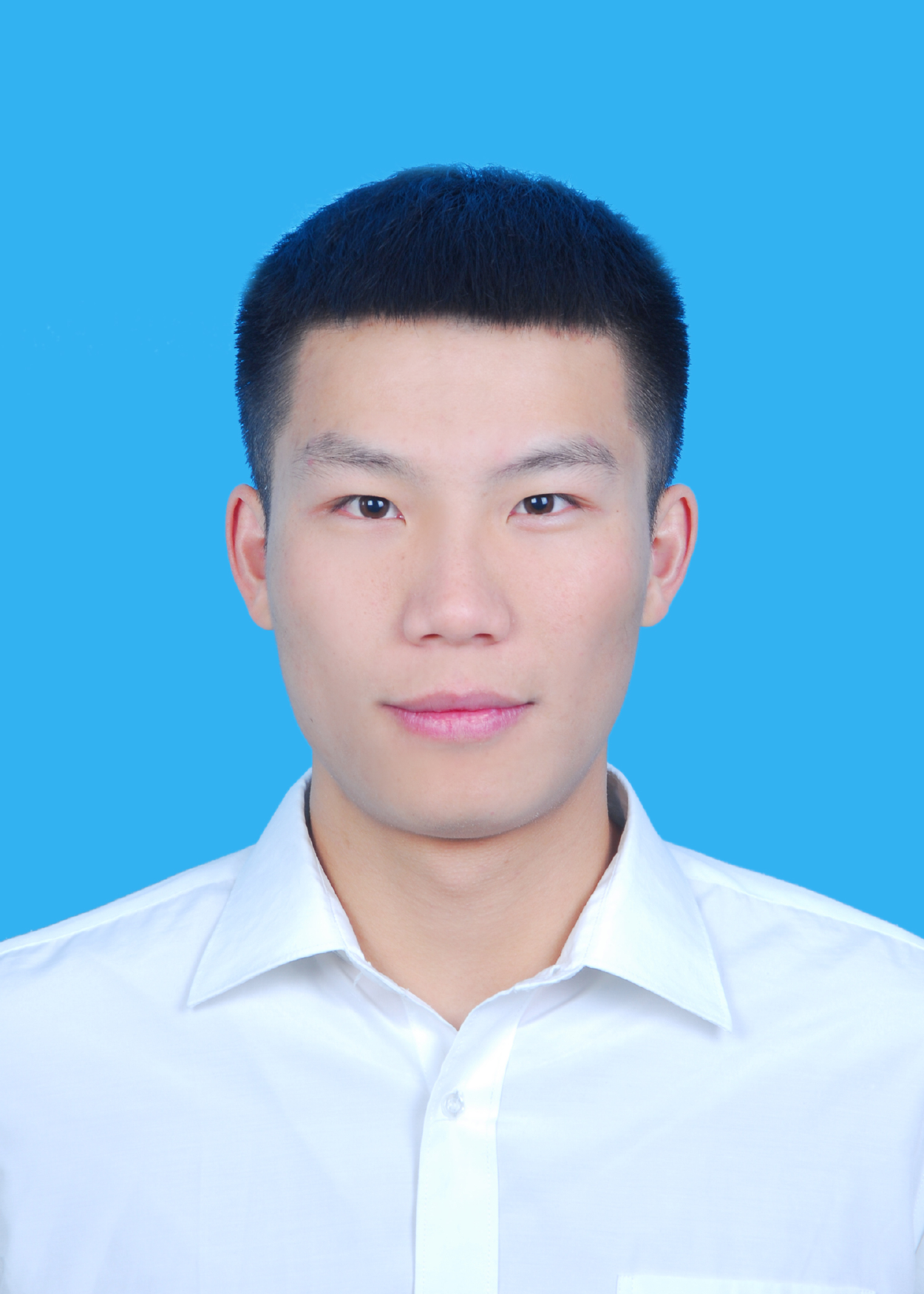}}]{Mingzhi Jiang}
  {Ming Zhi Jiang} received the B.E. degree in electronics information engineering from YanShan University (YSU), China, in 2019. He is currently pursuing the M.S. degree with the School of Artificial Intelligence, Beijing University of Posts and Telecommunications (BUPT), Beijing, China. His current work focuses on the optimization of collaborative intelligent transportation systems and the application of artificial intelligence. His research interests include edge computing, deep learning, and reinforcement learning.
\end{IEEEbiography}

\begin{IEEEbiography}[{\includegraphics[width=1in,height=1.25in,clip,keepaspectratio]{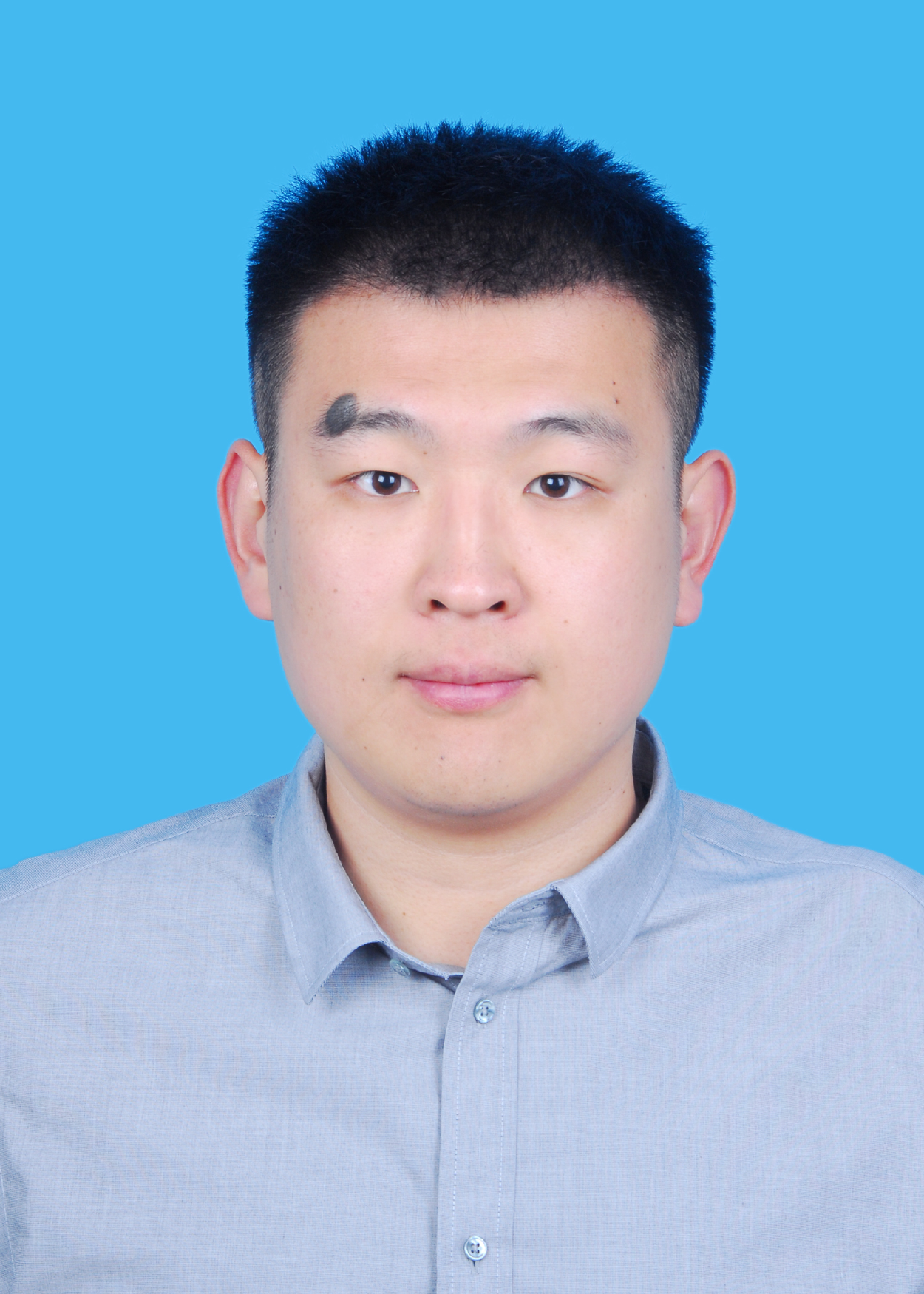}}]{Tianhao Wu}
  received the B.E. degree from Harbin Institute of Technology (HIT), China, in 2015. He is pursuing the Ph.D. degree in Information and Communication Engineering, Beijing University of Posts and Telecommunications (BUPT), China. His research interests include cooperative intelligent transportation systems and reinforcement learning.
\end{IEEEbiography}


\begin{IEEEbiography}[{\includegraphics[width=1in,height=1.25in,clip,keepaspectratio]{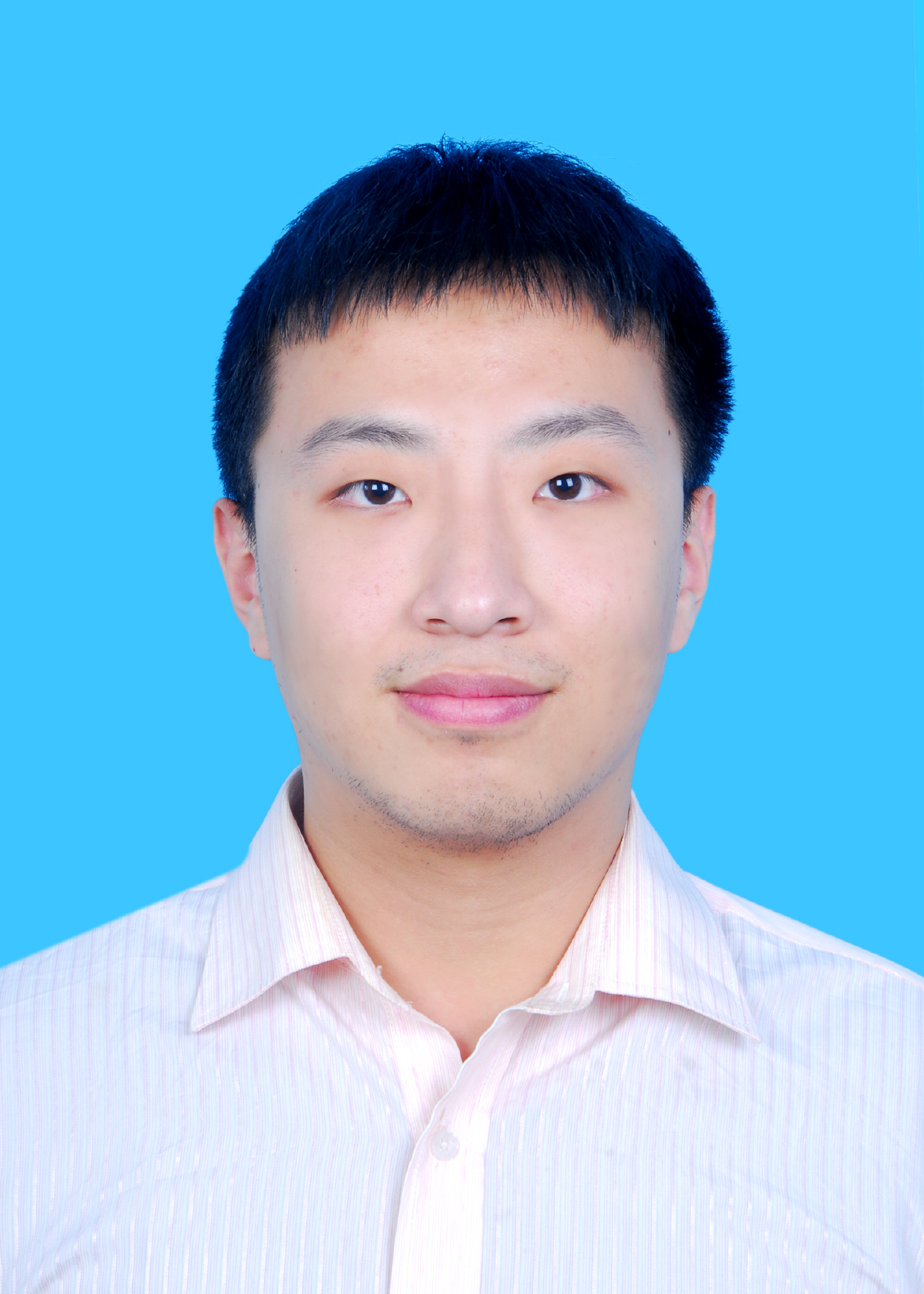}}]{Zhe Wang}
  received the B.E. degree in electronics and information engineering from Xi’an University of Posts and Telecommunications (XUPT), Xi’an, China, in 2019. He is currently pursuing the M.S. degree with the School of Artificial Intelligence, Beijing University of Posts and Telecommunications (BUPT), Beijing, China. His research interests are operation research in traffic congestion management, traffic assignment, and signal timing optimization.
\end{IEEEbiography}

\begin{IEEEbiography}[{\includegraphics[width=1in,height=1.25in,clip,keepaspectratio]{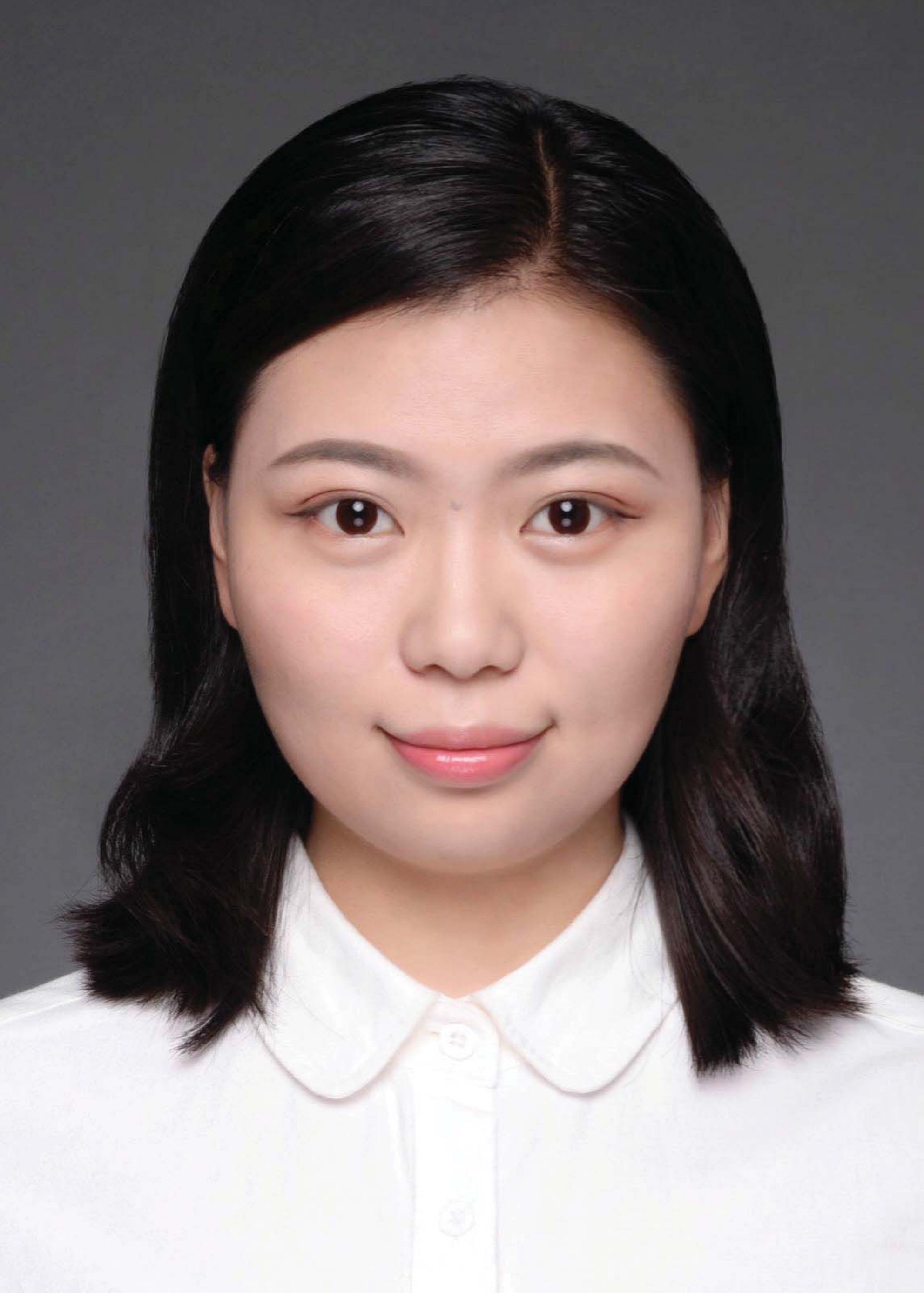}}]{Yi Gong}
  received the B.E. degree in information and engineering from the Xi'an University of Posts and Telecommunications (XUPT), Xi'an, Shannxi province, China. She received the M.S. and Ph.D. degrees in Information and Telecommunication Engineering, Beijing University of Posts and Telecommunications (BUPT), Beijing, China, in 2016 and 2020, respectively. She is currently a dual Ph.D. student in the Global Big Data Technologies Centre, University of Technology (UTS), Sydney, Australia. She is currently a lecture with the telecommunication and engineering, Beijing Information Science and Technology University (BISTU). Her research interests include channel modelling in vehicle networks, massive MIMO, low power communication, non-linear MIMO.
\end{IEEEbiography}

\begin{IEEEbiography}[{\includegraphics[width=1in,height=1.25in,clip,keepaspectratio]{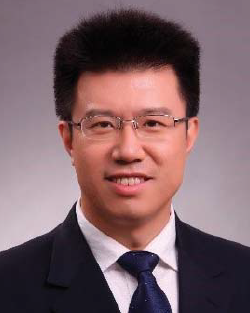}}]{Lin Zhang}
  received the B.S. and Ph.D. degrees from the Beijing University of Posts and Telecommunications (BUPT), Beijing, China, in 1996 and 2001, respectively. He is the vice chancellor of Beijing Information Science and Technology University (BISTU). He was a Postdoctoral Researcher with the Information and Communications University, Daejeon, Korea, from December 2000 to December 2002. He went to Singapore and held a Research Fellow position with Nanyang Technological University, Singapore, from January 2003 to June 2004. He joined BUPT in 2004 as a Lecturer, then an Associate Professor in 2005, and a Professor in 2011. He served the BUPT as the Director of Faculty Development Center, the Deputy Dean of Graduate School and the Dean of School of Information and Communication Engineering. He has authored more than 120 papers in referenced journals and international conferences. His research interests include mobile cloud computing and Internet of Things. 
\end{IEEEbiography}

\begin{IEEEbiography}[{\includegraphics[width=1in,height=1.25in,clip,keepaspectratio]{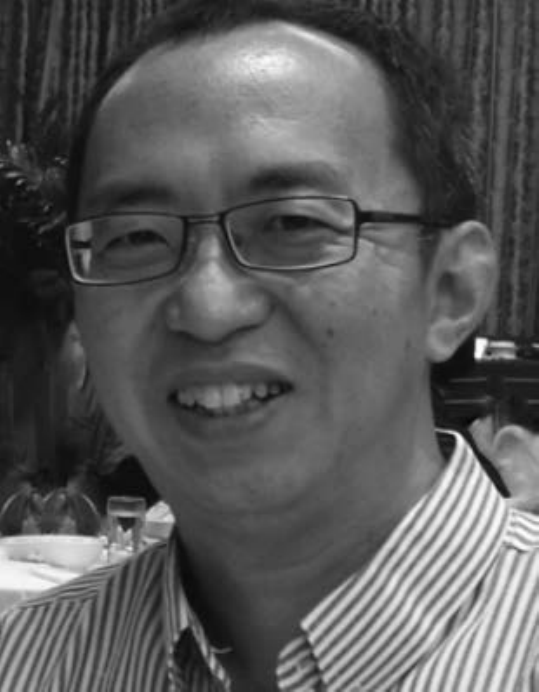}}]{Ren Ping Liu}
  received the B.E. and M.E. degrees from the Beijing University of Posts and Telecommunications, China, and the Ph.D. degree from the University of Newcastle, Australia.
	He is currently a Professor and the Head of the Discipline of Network and Cybersecurity, University of Technology Sydney. He is also the Co-Founder and CTO of Ultimo Digital Technologies Pty Ltd, developing IoT and Blockchain. Prior to that, he was a Principal Scientist and Research Leader with CSIRO, where he led wireless networking research activities. He specializes in system design and modeling and has delivered networking solutions to a number of government agencies and industry customers. His research interests include wireless networking, cybersecurity, and blockchain.
	Dr. Liu was the Founding Chair of the IEEE NSW VTS Chapter. He served as the technical program committee chair and the organizing committee chair in a number of IEEE conferences. He was a recipient of the Australian Engineering Innovation Award and the CSIRO Chairman Medal. He has over 150 research publications and has supervised over 30 Ph.D. students.
\end{IEEEbiography}




\end{document}